\documentclass[letterpaper]{article} 
\usepackage{main}  
\nocopyright
\usepackage[hyphens]{url}  
\usepackage{multirow}
\usepackage{tabularx}
\usepackage{amsmath}
\usepackage{amssymb}
\usepackage{pifont}
\usepackage{graphicx} 
\usepackage{natbib}  
\usepackage{caption} 
\usepackage{algorithm}
\usepackage{algorithmic}

\usepackage{newfloat}
\usepackage{listings}
\DeclareCaptionStyle{ruled}{labelfont=normalfont,labelsep=colon,strut=off} 
\floatstyle{ruled}
\newfloat{listing}{tb}{lst}{}
\floatname{listing}{Listing}

\usepackage{booktabs}

\title{Future-distilled Spectral Enhancement for Long-term Time Series Forecasting}
\author{
    Jingjing Bai\textsuperscript{\rm 1,3},
    Hiroshi Morioka\textsuperscript{\rm 1,2,3},
    Yoshinobu Kawahara\textsuperscript{\rm 1,3}
}
\affiliations{
    \textsuperscript{\rm 1}The University of Osaka, 
    \textsuperscript{\rm 2}Shiga University,
    \textsuperscript{\rm 3}RIKEN
}

\begin{document}

\maketitle

\begin{abstract}
Long-term time series forecasting models often suffer from over-smoothed predictions, especially at long horizons, where local fluctuations and informative mid- to high-frequency components are suppressed. Existing fixes either typically redesign the forecasting backbone or try to address distribution shifts in the time domain via plug-in normalization. We propose Future-distilled Spectral Enhancement (FSE), a lightweight, model-agnostic post-hoc framework that refines prediction from a pre-trained forecaster by correcting the spectral bias. During training, a teacher branch encodes the ground-truth future spectrum as privileged information, while a student branch learns to infer a corresponding latent representation from the historical input alone. To transfer this training-only future knowledge without explicit negative samples, we derive a non-contrastive dual-alignment objective that combines matched-pair cosine alignment with an EMA-based marginal regularizer. At inference time, the teacher branch is removed, and the student representation guides a bounded spectral correction network applied to the baseline prediction. Experiments on seven benchmarks and six forecasting backbones show that FSE improves accuracy in most settings, reduces high-frequency errors, and adds only small inference overhead. These results support future-distilled post-hoc correction as a practical way to recover predictable spectral residuals in long-term forecasting.
\end{abstract}

\begin{links}
     \link{Code}{https://github.com/Akira-221/FSE}
\end{links}

\section{Introduction}

Long-term time series forecasting is a fundamental task in sequential modeling, with broad applications in energy \cite{bilal2022comparative}, healthcare \cite{alaa2019attentive}, weather \cite{duchon2012time}, finance \cite{ariyo2014stock}, and industrial monitoring \cite{machines12060380}. Recent studies have advanced this field through decomposition-based models \cite{wu2021autoformer, zhou2022fedformer}, linear baselines \cite{zeng2023transformers}, patch-based Transformers \cite{nie2022time}, inverted attention \cite{liu2023itransformer}, multi-scale mixing \cite{wang2024timemixerdecomposablemultiscalemixing}, and frequency-domain architectures \cite{xu2023fits, yi2024frequency}. Yet, many models often suffer from over-smoothed predictions at long horizons. While dominant trends and periodic patterns can usually be captured, fine-grained temporal variations and informative spectral components are frequently weakened. 

\begin{figure}
    \centering
    \includegraphics[width=\linewidth]{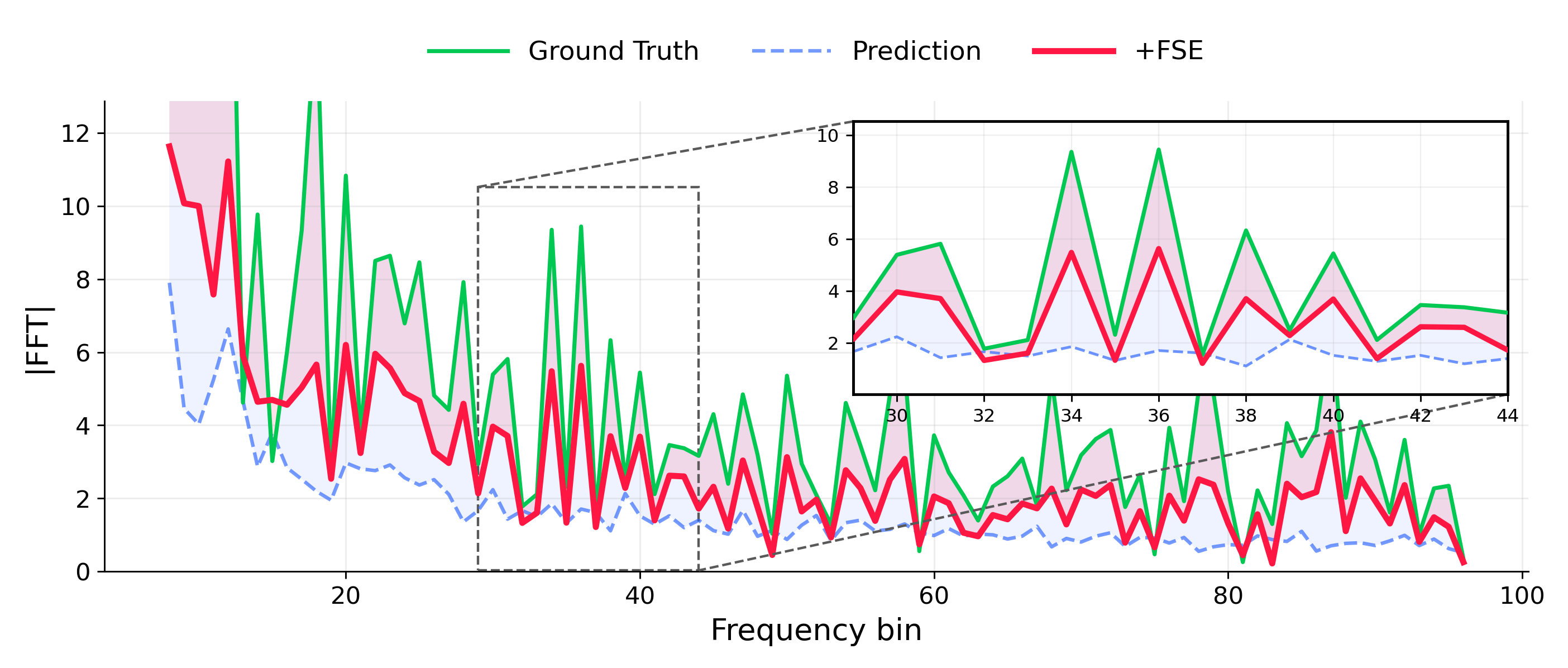}
    \caption{Frequency spectrum of ETTh1 dataset before and after FSE enhancement, using TimesNet as Baseline.}
    \label{fig:1}
\end{figure}

From the frequency-domain perspective, this phenomenon indicates a spectral mismatch between the predicted and true future signals (Figure~\ref{fig:1}). Point-wise losses such as MSE and MAE favor average trajectories under uncertainty, and the distribution shift between the lookback window and the forecast horizon further limits a model's ability to recover future-specific dynamics~\cite{kim2021reversible,liu2023nonstationarytransformersexploringstationarity,fan2023dishtsgeneralparadigmalleviating}. The result is a systematic gap between the predicted and true spectra, not merely a lack of model capacity. Rather than redesigning an increasingly complex backbone, we formulate the problem as post hoc spectral refinement: given a pretrained forecasting model, we aim to recover the predictable spectral residual between its prediction and the ground-truth future without modifying the original forecasting backbone.

To achieve this goal, we propose \textit{Future-distilled Spectral Enhancement} (FSE) framework, a lightweight, model-agnostic module trained in two stages. First, the baseline forecaster is trained and then frozen; its output serves as a fixed coarse prediction. Second, FSE is trained to correct this baseline in the frequency domain. During the second stage, a future teacher encoder extracts spectral information from the ground-truth future, while a past student encoder learns to infer future-aware representations only from historical inputs. At inference time, the teacher branch is removed, and the student representation guides a spectral correction network to refine the baseline forecast. 

The future guidance is achieved through latent alignment inspired by representation learning. Instead of explicit contrastive learning~\cite{oord2018representation, chen2020simple}, which may introduce false negative pairs due to shared seasonal and periodic structures among different future segments, FSE adopts a non-contrastive alignment objective derived from a mutual-information-based contrastive formulation. The objective combines a sample-wise term that aligns each student representation with its matched future teacher representation and an EMA-based marginal regularizer that controls its similarity to the overall future representation distribution without explicitly constructing cross-sample negative pairs, enabling the student representation to preserve informative future spectral characteristics without requiring future access during inference.

At inference, the teacher is discarded, and the student representation conditions a lightweight correction network that adjusts the baseline spectrum with a bounded, sample-dependent magnitude before reconstructing the final prediction.

Extensive experiments on seven benchmarks with six forecasting backbones demonstrate that FSE consistently improves forecasting accuracy, reduces spectral errors, and introduces only limited computational overhead. Further frequency-domain analyses verify that the proposed method effectively recovers missing spectral components while preserving the original forecasting structure.

Our contributions are summarized as follows:
\begin{itemize}
    \item We introduce a post-hoc spectral refinement perspective for long-term forecasting, where forecast errors are interpreted as predictable spectral residuals rather than requiring a new forecasting backbone.

    \item We propose FSE, a future-distillation framework that transfers future spectral information into a past-conditioned latent representation through non-contrastive alignment.

    \item We develop a spectral correction network guided by the distilled representation to recover attenuated frequency components from frozen baseline predictions.
    
    \item Extensive experiments across datasets and forecasting architectures demonstrate the effectiveness, robustness, and efficiency of FSE.
\end{itemize}

\section{Related Work}

\paragraph{Frequency-domain forecasting.}
Frequency-domain modeling has become an important way for long-term time series forecasting, since periodicity and multi-scale variation are always compact in the spectral space. Representative methods incorporate Fourier or frequency-aware operations into forecasting backbones, including FEDformer~\cite{zhou2022fedformer}, FiLM~\cite{zhou2022film}, FITS~\cite{xu2023fits}, FreTS~\cite{yi2024frequency}, FTMixer~\cite{li2024ftmixer}, Koopa~\cite{liu2023koopa}, and FreEformer~\cite{yue2025freeformer}. Other architectures, such as TimesNet~\cite{wu2022timesnet} and TimeMixer~\cite{wang2024timemixerdecomposablemultiscalemixing}, also exploit periodic or multi-scale structures closely related to frequency analysis. These methods mainly redesign the backbone to generate the forecast in the frequency or time-frequency domain. In contrast, FSE treats an arbitrary forecaster's output as a coarse prediction and learns to correct the predictable spectral residuals that are weakened by over-smoothing.

\paragraph{Post-hoc correction and adaptive normalization.}
A related line of work improves forecasting robustness through plug-in correction or normalization modules. RevIN~\cite{kim2021reversible}, DAIN~\cite{passalis2019deepadaptiveinputnormalization}, Non-stationary Transformers~\cite{liu2023nonstationarytransformersexploringstationarity}, Dish-TS~\cite{fan2023dishtsgeneralparadigmalleviating}, GAS-Norm~\cite{Urettini_2024}, NoRIN~\cite{zhang2026norinbackboneadaptivereversiblenormalization}, and frequency-aware variants such as FAN~\cite{ye2024frequencyadaptivenormalizationnonstationary} and FredNormer~\cite{piao2024frednormerfrequencydomainnormalization}. These mainly operate as input/output normalization or distribution adjustment mechanisms. FSE instead corrects the baseline forecast directly in the spectral domain, using future-distilled guidance to recover attenuated frequency components rather than rescaling, stationarizing, or reweighting the observed sequence.

\paragraph{Time-series representation learning.}
Contrastive and self-supervised representation learning has also been widely explored for time series, including TS2Vec~\cite{yue2022ts2vecuniversalrepresentationtime}, CoST~\cite{woo2022costcontrastivelearningdisentangled}, TF-C~\cite{zhang2022selfsupervisedcontrastivepretrainingtime}, and SimTS~\cite{zheng2024simplecontrastiverepresentationlearning}. These works show that representation learning can improve temporal modeling, but explicit negative sampling may be unreliable in forecasting since different samples often share similar trends, cycles, or spectral patterns. FSE adopts a related past-future alignment perspective, but converts the contrastive objective into a non-contrastive dual-alignment loss, making the learned representation directly serve spectral forecast correction.

\section{Preliminary}\label{sec:preliminary}
\begin{figure*}
    \centering
    \includegraphics[width=0.9\linewidth]{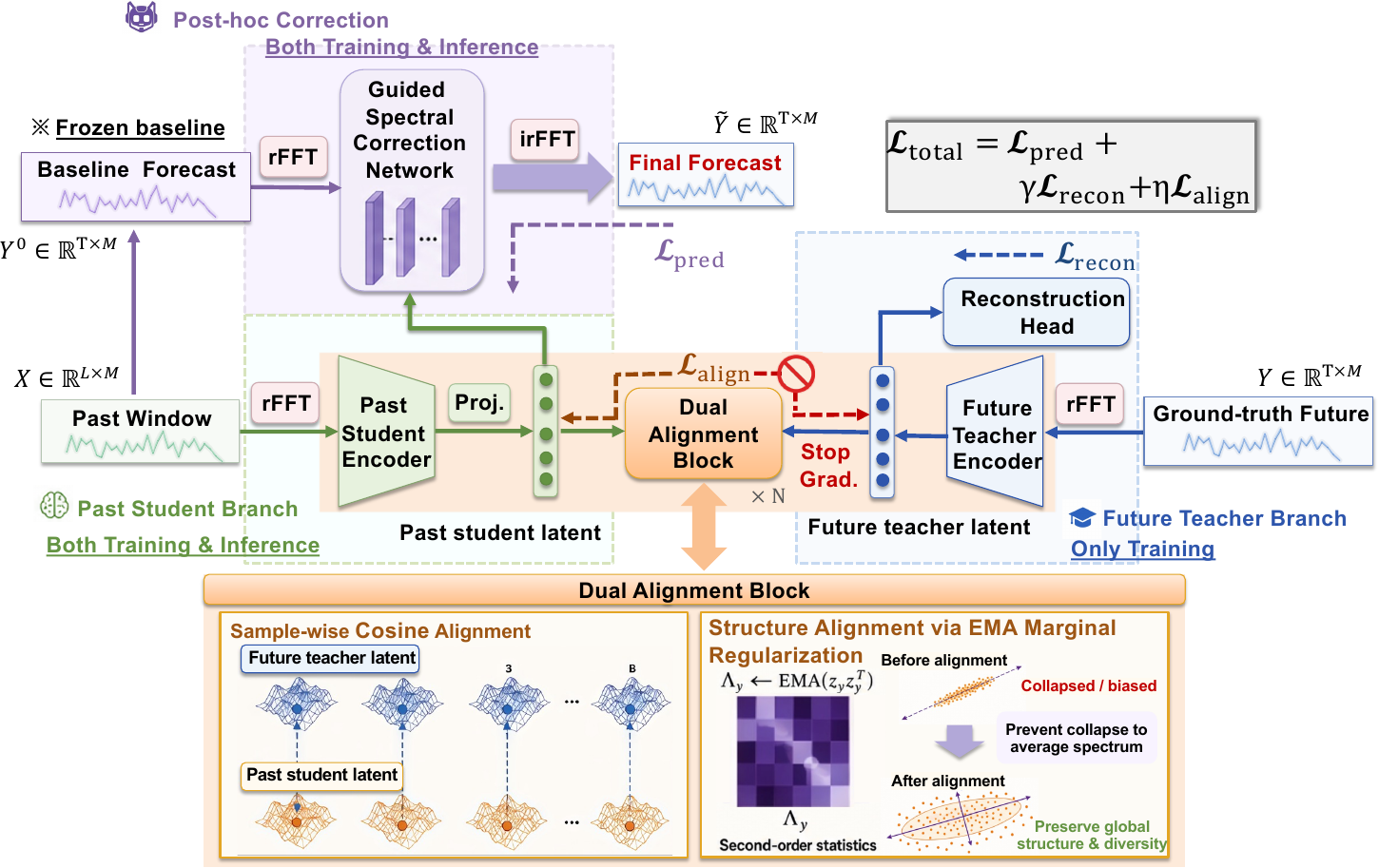}
    \caption{Overview of proposed FSE framework.}
    \label{fig:overview}
\end{figure*}
Let \(X\) and \(Y\) denote the historical window and its future, with joint distribution \(p(x,y)\) and marginals \(p(x)\) and \(p(y)\). Their statistical dependency is measured by the \(f\)-mutual information
\[
I_f(X,Y)
=
\int
f\left(
\frac{p(x,y)}{p(x)p(y)}
\right)
p(x)p(y)\,dx\,dy,
\]
where \(f:\mathbb{R}_{+}\rightarrow\mathbb{R}\) is convex with \(f(1)=0\).


By the variational representation of \(f\)-divergence~\cite{nachum2020reinforcement, nguyen2010estimating},
\begin{align*}
I_f&(X,Y)
=\\
&\sup_{t:\mathcal{X}\times\mathcal{Y}\rightarrow\mathbb{R}}
\left\{
\mathbb{E}_{p(x,y)}[t(x,y)]
-
\mathbb{E}_{p(x)p(y)}
\left[
f^{*}(t(x,y))
\right]
\right\},
\end{align*}
with \(f^{*}\) the Fenchel conjugate of \(f\). The first term rewards agreement between matched pairs; the second regularizes agreement under independent pairing. This formulation therefore generalizes the central principle of contrastive learning: preserving the dependency between paired views while controlling their similarity under independent pairing.

Parameterizing \(t(x,y)\) through representations \(z_x=\phi_x(x),z_y=\phi_y(y)\), different choices of \(f\) and \(t\) recover different contrastive objectives, including InfoNCE~\cite{oord2018representation}, SimCLR \cite{chen2020simple}, and spectral contrastive learning \cite{haochen2021provable}.

Two issues make this objective impractical for time series forecasting. First, estimating the marginal term over all cross-sample pairs scales quadratically with batch size. Second, futures from different samples often share trend, seasonality, or periodicity, so treating them as negatives risks false-negative supervision. We resolve both by converting \(I_f(X,Y)\) into a non-contrastive dual-alignment objective, instantiated below with the Pearson \(\chi^2\) divergence and generalized to the \(\alpha\)-divergence family in the Appendix.

\section{Proposed Method}

In this section, we present the proposed \textit{Future-distilled Spectral Enhancement} (FSE) framework, as illustrated in Figure~\ref{fig:overview}. FSE is a lightweight and model-agnostic framework for correcting the spectral bias of an existing forecaster. Instead of replacing or redesigning the forecasting backbone, FSE refines its initial prediction through future-informed representation learning and guided spectral correction. We first formulate the post-hoc forecast correction problem and then introduce the main components of the proposed framework.

\subsection{Problem Formulation}

Consider a multivariate time series forecasting task. Let \(X=\{x_i\}_{i=1}^{M}\in\mathbb{R}^{L\times M}\) denote the historical input, where \(L\) is the look-back window and \(M\) is the number of variables. The corresponding future target is
defined as \( Y=\{y_i\}_{i=1}^{M}\in\mathbb{R}^{T\times M}\), where \(T\) denotes the forecasting horizon. A forecasting model generally learns a mapping from the historical window to the future sequence.

In this work, we consider a post-hoc correction setting. Given a baseline forecaster, its initial prediction is \(Y^{0}\in\mathbb{R}^{T\times M}\). The baseline is pretrained and then frozen before training FSE, allowing the proposed framework to be applied to different forecasting architectures without retraining them from scratch.

Although \(Y^{0}\) typically captures the dominant trend and periodic structure, it tends to attenuate local variations and mid- to high-frequency components. FSE learns a correction function \(\widetilde{Y}=\mathcal{R}_{\theta}(X,Y^{0})\) that refines \(Y^0\) in the frequency domain. Concretely, \(Y^0\) and \(Y\) are transformed via the real-valued FFT into complex spectra
\[
F^0_y=\operatorname{rFFT(Y^0)},\: F_y=\operatorname{rFFT(Y)\in\mathbb{C}^{K_y\times M}},
\]
where \(K_y=\left\lfloor\frac{T}{2}\right\rfloor+1\), and \(\mathcal{R}_{\theta}\) operates on \(F^0_y\) rather than on \(Y^0\) directly. The refined forecast is obtained from refind spectrum $\widetilde{F}_y$ via an inverse real-valued FFT transform, \(\widetilde{Y}=\operatorname{irFFT(\widetilde{F}_y, T)}\).

During training, the ground-truth future \(Y\) is additionally available. FSE treats it as privileged teacher information to identify spectral structures that are missing from the baseline prediction. Such future information is distilled into a past-conditioned representation and is not directly accessed during inference. The objective is therefore not to reconstruct the entire future from scratch, but to recover the predictable spectral residuals omitted by the baseline while preserving its reliable forecasting components.

\subsection{Framework Overview}

Training proceeds in two stages. The baseline forecaster is trained and frozen first; FSE is then trained on top of it in the second stage, as illustrated in Figure~\ref{fig:overview}. Within this stage, a future teacher branch and a past student branch share encoder parameters, encouraging both to learn a common spectral representation. The teacher branch is trained only in this stage, on the ground truth future \(Y\), while the student branch, used in both training and inference, takes the history input \(X\). A dual-alignment objective couples them at every layer, transferring future spectral knowledge into the student representation without requiring access to \(Y\) at inference. Then, a guided spectral correction network consumes the student representation to refine \(F^0_y\).

\subsection{Future Teacher Branch}

The future teacher branch is used only during training to extract future-specific spectral information from \(Y\). Its output serves as an informative target for distilling spectral patterns that may be attenuated or omitted by the baseline forecast.

Since \(F_y\) is complex-valued while the encoder operates with standard real-valued layers, we first map it to a real feature by concatenating its real and imaginary parts along the channel dimension:
\[
\psi(F_y)=\text{Concat}(\text{Re}(F_y),\text{Im}(F_y))\in\mathbb{R}^{K_y\times D}, \quad D=2M
\]
Setting \(z_y^0=\psi(F_y)\), the teacher representation is computed by an \(N\)-layer residual encoder:
\begin{align*}
    z_y^n=z_y^{n-1}+\text{Linear}(\alpha(\text{Linear}(z_y^{n-1}))),\quad n=1,\cdots,N
\end{align*}
with \(\alpha\) a nonlinear activation. To keep \(z_y^N\) informative rather than collapsing to an arbitrary alignment target, we attach a reconstruction head \(\widehat{S}_{recon}=\text{Linear}(z_y^N)\) trained to regress \(\psi(F_y)\), optimized by frequency domain MAE \(\mathcal{L}_{\mathrm{recon}}\). 

\subsection{Past Student Branch}
The past student branch produces future-aware guidance using only the historical input. Given \(X\in\mathbb{R}^{L\times M}\), its initial spectral representation is defined as \(\widetilde{z}_x^0=\psi\left(\operatorname{rFFT}(X)\right)\in\mathbb{R}^{K_x\times D}, \:K_x=\left\lfloor L/2\right\rfloor+1\). The student and teacher branches share the parameters of the \(N\)-layer residual encoder. Accordingly, for \(n\)-th layer:
\begin{align*}
    \tilde{z}_x^n=\tilde{z}_x^{n-1}+\text{Linear}(\alpha(\text{Linear}(\tilde{z}_x^{n-1}))),\quad n=1,\cdots,N
\end{align*}

When \(L\neq T\) in general, \(K_x\neq K_y\), and the two branches produce representations over different numbers of frequency tokens. We therefore add a per-layer projection
\[
    z_x^n=P_x^n\left(\widetilde{z}_x^n\right),
    \quad
    P_x^n:
    \mathbb{R}^{K_x\times D}
    \rightarrow
    \mathbb{R}^{K_y\times D},
\]
This additional projection layer, implemented as a lightweight Linear layer, maps the historical representation into the same token space as \(z_y^n\), enabling layer-wise alignment between the two branches. Beyond matching token dimensions, this projection interposes an additional trainable transformation between the past encoder and the alignment objective, so that alignment gradients reach the encoder through a learned map rather than acting on its output directly.

During training, \(z_x^n\) is aligned with its corresponding future representation \(z_y^n\) at every layer through the dual-alignment objective described next. At inference, the future teacher branch and reconstruction head are removed. Only the student branch is retained to provide future-aware spectral guidance without access to the ground-truth future.

\subsection{Dual Alignment Block}
\label{sec:dual_alignment}
The future and past branches encode \(X\) and \(Y\) independently, but their latent spaces still diverge due to the distribution shift between history and future. We align them at each layer using the \(f\)-MI objective from the Preliminary, instantiated with the Pearson \(\chi^2\) divergence \(f(u)=(u-1)^2\), whose conjugate is \(f^*(t)=t^2/4+t\). Taking \(t(x,y)=2(\phi(x)^\top\phi(y)-1)\) and dropping constants gives
\[
\max_{\phi_x,\phi_y}
2\mathbb{E}_{p(x,y)}
\left[
\phi_x(x)^\top\phi_y(y)
\right]
-
\mathbb{E}_{p(x)p(y)}
\left[
\phi_x(x)^\top\phi_y(y)
\right]^2
\]
The first term is the matched-pair dependency term; the second is the marginal term that would otherwise require explicit negative pairs. We reformulate each without cross-sample negatives.

\paragraph{Sample-wise Cosine Alignment.}

For layer \(n\), the matched-pair term is estimated over a mini-batch of size \(B\) as
\[
    \mathbb{E}_{p(x,y)}
    \left[
        \phi_x(x)^\top\phi_y(y)
    \right]
    =
    \frac{1}{B}
    \sum_{i=1}^{B}
    z_{x,i}^{n\top}z_{y,i}^{n}.
\]

Maximizing it is equivalent to minimizing
\[
    \mathcal{L}_{\mathrm{cos}}^{n}
    =
    \frac{1}{B}
    \sum_{i=1}^{B}
    \left(
        1-
        z_{x,i}^{n\top}
        \operatorname{sg}(z_{y,i}^{n})
    \right),
\]
where \(\operatorname{sg}(\cdot)\) indicates stop-gradient operation applied only on the teacher side~\cite{chen2021exploring, hu2026bridgingpastfuturedistributionaware}. The asymmetry is deliberate, since distillation is intended to move \(z_x\) toward \(z_y\), not to draw the two representations toward each other. Using this stop gradient ensures that gradients from the alignment loss are
only propagated to the past student branch, while the future representation does not drift toward the past.

\paragraph{Structure Alignment via EMA Marginal Regularization.}

The product-of-marginals term requires independently paired past and future representations. Its empirical estimate is
\[
    \mathbb{E}_{p(x)p(y)}
    \left[
            \phi_{x}(x)^{\top}\phi_{y}(y)
    \right]^2
    =
    \frac{1}{B^2}
    \sum_{i=1}^{B}
    \sum_{j=1}^{B}
    \left(
        z_{x,i}^{n\top}z_{y,j}^{n}
    \right)^2,
\]
which explicitly involves all cross-sample pairs. It can be expanded as
\[
    \frac{1}{B}
    \sum_{i=1}^{B}
    z_{x,i}^{n\top}
    C_y^{n}
    z_{x,i}^{n},
\]
where
\[
    C_y^{n}
    =
    \frac{1}{B}
    \sum_{j=1}^{B}
    z_{y,j}^{n}z_{y,j}^{n\top}
\]
is the batch second-moment matrix of the future representations.

To reduce the dependence on individual mini-batches and approximate the global future marginal distribution, we maintain an exponential moving average (EMA) statistic for each encoder layer:
\[
    \Lambda_y^{n,(t)}
    =
    \beta\Lambda_y^{n,(t-1)}
    +
    (1-\beta)\operatorname{sg}
    \left(C_y^{n,(t)}\right),
\]
where \(t\) is the optimization step and \(\beta\in[0,1)\) is the EMA decay factor (set as 0.8). We then define the marginal regularization term as
\[
    \mathcal{L}_{\mathrm{ema}}^{n}
    =
    \frac{1}{B}
    \sum_{i=1}^{B}
    z_{x,i}^{n\top}
    \operatorname{sg}(\Lambda_y^{n})
    z_{x,i}^{n}.
\]

Thus, \(\mathcal{L}_{\mathrm{ema}}^{n}\) approximates the squared similarity to the marginal future representation distribution without explicitly enumerating negative pairs. It prevents the student from becoming uniformly similar to the overall teacher
distribution while preserving its agreement with the matched teacher.

\paragraph{Dual Alignment Objective.}

\begin{table*}[t]
\centering
\setlength{\tabcolsep}{1.2pt}
\begin{tabular}{cc|ccc|ccc|ccc|ccc|ccc|ccc}
\toprule

\multicolumn{2}{c|}{\textbf{Models}}
& \multicolumn{3}{c|}{\textbf{TimesNet}}
& \multicolumn{3}{c|}{\textbf{TimeMixer}}
& \multicolumn{3}{c|}{\textbf{Dualformer}}
& \multicolumn{3}{c|}{\textbf{DLinear}}
& \multicolumn{3}{c|}{\textbf{PatchTST}}
& \multicolumn{3}{c}{\textbf{iTransformer}}
\\

\cmidrule(lr){1-2}\cmidrule(lr){3-5} \cmidrule(lr){6-8} \cmidrule(lr){9-11} \cmidrule(lr){12-14} \cmidrule(lr){15-17} \cmidrule(lr){18-20}

\multicolumn{2}{c|}{Dataset $\quad$Metric}
& Base & +FSE & Imp.
& Base & +FSE & Imp.
& Base & +FSE & Imp.
& Base & +FSE & Imp.
& Base & +FSE & Imp.
& Base & +FSE & Imp.
\\

\midrule

\multirow{2}{*}{\emph{ETTh1}}
& MSE
& 0.458 & \textbf{0.408} & \textbf{10.9}
& 0.411 & \textbf{0.401} & \textbf{2.4}
& 0.407 & \textbf{0.396} & \textbf{2.7}
& 0.439 & \textbf{0.430} & \textbf{2.1}
& 0.418 & \textbf{0.409} & \textbf{2.2}
& 0.445 & \textbf{0.421} & \textbf{5.4}
\\

& MAE
& 0.450 & \textbf{0.427} & \textbf{5.1}
& 0.423 & \textbf{0.418} & \textbf{1.2}
& 0.421 & \textbf{0.420} & \textbf{0.2}
& 0.449 & \textbf{0.445} & \textbf{0.9}
& 0.437 & \textbf{0.418} & \textbf{4.3}
& 0.434 & \textbf{0.432} & \textbf{0.5}
\\

\midrule

\multirow{2}{*}{\emph{ETTh2}}
& MSE
& 0.414 & \textbf{0.336} & \textbf{18.8}
& 0.316 & 0.333 & -5.4
& 0.335 & \textbf{0.324} & \textbf{3.3}
& 0.458 & \textbf{0.445} & \textbf{2.8}
& 0.343 & \textbf{0.340} & \textbf{0.9}
& 0.374 & \textbf{0.345} & \textbf{7.8}
\\

& MAE
& 0.427 & \textbf{0.387} & \textbf{9.4}
& 0.384 & \textbf{0.381} & \textbf{0.8}
& 0.377 & \textbf{0.373} & \textbf{1.1}
& 0.459 & \textbf{0.443} & \textbf{3.5}
& 0.388 & 0.388 & --
& 0.399 & \textbf{0.388} & \textbf{2.8}
\\

\midrule

\multirow{2}{*}{\emph{ETTm1}}
& MSE
& 0.388 & \textbf{0.350} & \textbf{9.8}
& 0.348 & \textbf{0.342} & \textbf{1.7}
& 0.345 & \textbf{0.339} & \textbf{1.7}
& 0.359 & \textbf{0.355} & \textbf{1.1}
& 0.350 & 0.351 & -0.3
& 0.407 & \textbf{0.369} & \textbf{9.3}
\\

& MAE
& 0.406 & \textbf{0.382} & \textbf{5.9}
& 0.376 & \textbf{0.367} & \textbf{2.4}
& 0.368 & \textbf{0.361} & \textbf{1.9}
& 0.380 & \textbf{0.375} & \textbf{1.3}
& 0.381 & 0.383 & -0.5
& 0.410 & \textbf{0.397} & \textbf{3.2}
\\

\midrule

\multirow{2}{*}{\emph{ETTm2}}
& MSE
& 0.291 & \textbf{0.262} & \textbf{10.0}
& 0.256 & \textbf{0.243} & \textbf{5.1}
& 0.245 & \textbf{0.238} & \textbf{2.9}
& 0.281 & \textbf{0.275} & \textbf{2.1}
& 0.250 & \textbf{0.242} & \textbf{3.2}
& 0.288 & \textbf{0.265} & \textbf{8.0}
\\

& MAE
& 0.333 & \textbf{0.321} & \textbf{3.6}
& 0.316 & \textbf{0.302} & \textbf{4.4}
& 0.304 & \textbf{0.292} & \textbf{3.9}
& 0.343 & \textbf{0.330} & \textbf{3.8}
& 0.314 & \textbf{0.310} & \textbf{1.3}
& 0.332 & \textbf{0.310} & \textbf{6.6}
\\

\midrule

\multirow{2}{*}{\emph{Electricity}}
& MSE
& 0.193 & \textbf{0.159} & \textbf{17.6}
& 0.165 & \textbf{0.154} & \textbf{6.7}
& 0.158 & \textbf{0.156} & \textbf{1.3}
& 0.167 & \textbf{0.164} & \textbf{1.8}
& 0.161 & \textbf{0.157} & \textbf{2.5}
& 0.161 & \textbf{0.152} & \textbf{5.6}
\\

& MAE
& 0.295 & \textbf{0.256} & \textbf{13.2}
& 0.253 & \textbf{0.244} & \textbf{3.6}
& 0.249 & \textbf{0.243} & \textbf{2.4}
& 0.264 & \textbf{0.260} & \textbf{1.5}
& 0.255 & \textbf{0.253} & \textbf{0.8}
& 0.257 & \textbf{0.250} & \textbf{2.7}
\\

\midrule

\multirow{2}{*}{\emph{Traffic}}
& MSE
& 0.620 & \textbf{0.404} & \textbf{34.8}
& 0.393 & \textbf{0.387} & \textbf{1.5}
& 0.406 & \textbf{0.402} & \textbf{1.0}
& 0.435 & 0.435 & --
& 0.407 & \textbf{0.400} & \textbf{1.7}
& 0.428 & \textbf{0.413} & \textbf{3.5}
\\

& MAE
& 0.336 & \textbf{0.281} & \textbf{16.4}
& 0.271 & \textbf{0.255} & \textbf{5.9}
& 0.268 & \textbf{0.258} & \textbf{3.7}
& 0.298 & \textbf{0.296} & \textbf{0.7}
& 0.282 & \textbf{0.278} & \textbf{1.4}
& 0.282 & \textbf{0.270} & \textbf{4.3}
\\

\midrule

\multirow{2}{*}{\emph{Solar}}
& MSE
& 0.244 & \textbf{0.210} & \textbf{13.9}
& 0.192 & 0.195 & -1.6
& 0.191 & 0.198 & -3.7
& 0.329 & \textbf{0.320} & \textbf{2.7}
& 0.256 & \textbf{0.223} & \textbf{12.9}
& 0.233 & \textbf{0.230} & \textbf{1.3}
\\

& MAE
& 0.334 & \textbf{0.248} & \textbf{25.7}
& 0.244 & \textbf{0.230} & \textbf{5.7}
& 0.240 & \textbf{0.238} & \textbf{0.8}
& 0.390 & \textbf{0.383} & \textbf{1.8}
& 0.298 & \textbf{0.269} & \textbf{9.7}
& 0.262 & \textbf{0.257} & \textbf{1.9}
\\

\bottomrule
\end{tabular}
\caption{Average performance comparison between original models and models enhanced with FSE across six forecasting backbones. Improvement (\%) is computed relative to the original model. Positive values indicate performance gains and negative values indicate degradation.}
\label{tab:fse_all_models}
\end{table*}

Combining the Sample-wise cosine alignment and EMA marginal regularization gives the layer-wise dual-alignment objective:
\[
    \mathcal{L}_{\mathrm{align}}
    =
    \frac{1}{N}\sum^N_{n=1}(\lambda_{\mathrm{cos}}\mathcal{L}_{\mathrm{cos}}^{n}
    +
    \lambda_{\mathrm{ema}}
    \mathcal{L}_{\mathrm{ema}}^{n}).
\]
The \(f\)-MI formulation determines that the objective comprises a matched-pair term and a marginal term, but not their relative weight. The ratio \(\lambda_{\mathrm{cos}}/\lambda_{\mathrm{ema}}=2\) is specific to the Pearson \(\chi^2\) instantiation rather than common to the broader \(\alpha\)-divergence family to which this marginal form generalizes (Appendix). We therefore treat \(\lambda_{\mathrm{cos}}, \lambda_{\mathrm{ema}}\) as tunable hyperparameters to account for differing empirical scales. Appendix Figure~\ref{cosine_weight_sensitivity} shows the optimum shifting mildly across datasets and prediction horizons while performance remains robust, with the Pearson \(\chi^2\) ratio falling within the near-optimal region. The result is non-contrastive, since no cross-sample negatives are constructed, while both terms remain tied to the joint and marginal components of \(I_f(X,Y)\). 

\subsection{Guided Spectral Correction Network}
The correction network takes the baseline spectrum \(F_y^0\) and the distilled student representation \(z_x\) to produce a bounded adjustment to \(F_y^0\). Since \(F_y^0\) is complex and \(z_x\) is real, we again work in the concatenated real representation, forming the correction context
\[
    g=
    \operatorname{Concat}(
        \psi(F_y^0),z_x)\in\mathbb{R}^{K_y\times (D+2M)}.
\]
A lightweight spectral correction network predicts a real-valued candidate correction
\begin{align*}
\Delta S_y^{\mathrm{raw}}
   & =
    G_{\phi}(g)\\
   & =
    \operatorname{Conv1D}_2
    \left(
        \operatorname{GELU}
        \left(
            \operatorname{Conv1D}_1(g)
        \right)
    \right)\in\mathbb{R}^{K_y\times 2M}    
\end{align*}
which is split along the channel dimension into real and imaginary halves and recombined into the complex candidate \(\Delta F_y^{\mathrm{raw}}\in\mathbb{C}^{K_y\times M}\)

An uncontrolled correction of this form could overwrite reliable parts of the baseline spectrum, so we bound its magnitude relative to \(F^0_y\) with an element-wise risk-control operator applied to the complex modulus:
\[
    \Theta
    =
    \min
    \left(
        1,
        \frac{\alpha\left|F_y^0\right|}
        {\left|\Delta F_y^{\mathrm{raw}}\right|+\epsilon}
    \right),\
    \Delta F_y
    =
    \Theta\odot\Delta F_y^{\mathrm{raw}},
\]
where \(\alpha\) specifies the maximum correction magnitude relative to the baseline spectrum and \(\epsilon\) is a small constant that ensures numerical stability. Accordingly, the refined spectrum and final forecast are
\[
    \widetilde{F}_y
    =
    F_y^0+\Delta F_y,
    \qquad
    \widetilde{Y}
    =
    \operatorname{irFFT}(\widetilde{F}_y,T).
\]

The correction network \(G_\phi\) determines the sample-dependent direction of correction,
while the risk-control operator \(\Theta\) bounds its magnitude relative to the
baseline spectral content. This design enables the network to rescale specific frequency components without the risk of amplifying them unboundedly or discarding reliable baseline content.

\subsection{Overall Objective}

FSE is trained end-to-end in this second stage with the baseline forecaster kept frozen, using
\[
\mathcal{L}_{\mathrm{total}}
=
\mathcal{L}_{\mathrm{pred}}
+
\gamma\mathcal{L}_{\mathrm{recon}}
+
\eta\mathcal{L}_{\mathrm{align}},
\]
where \(\mathcal{L}_{\mathrm{pred}}\) is the time-domain MAE between \(\widetilde{Y}\) and \(Y\), \(\mathcal{L}_{\mathrm{recon}}\) is the frequency domain MAE between \(\widehat{S}_{recon}\) and \(\psi(F_y)\), and \(\mathcal{L}_{\mathrm{align}}\) is the dual-alignment loss from the previous section.

\section{Experiments}

In this section, we conducted extensive experiments on widely used benchmark datasets, comparing our proposed FSE framework with mainstream forecasting models under a multivariate long-term forecasting setting to demonstrate its effectiveness.
\subsection{Experimental Setup}
\paragraph{Datasets.} We evaluated on seven real-world multivariate benchmarks: Electricity, Solar Energy, Traffic, and four ETT datasets (ETTh1, ETTh2, ETTm1, ETTm2). Dataset statistics, including sampling frequency, number of variables, sequence length, and time ranges, are summarized in Appendix Table~\ref{tab:data}. Following standard protocol~\cite{wu2021autoformer}, we split each dataset into training, validation, and test sets with ratios of 6:2:2 for the ETT datasets and 7:1:2 for the rest.

\paragraph{Baseline Models.} FSE is designed as a model-agnostic plugin. To demonstrate its effectiveness, we selected six mainstream baseline models based on diverse architectures: \textbf{TimesNet} \cite{wu2022timesnet}, \textbf{TimeMixer} \cite{wang2024timemixerdecomposablemultiscalemixing}, \textbf{Dualformer} \cite{bai2026dualformer}, \textbf{DLinear} \cite{zeng2023transformers}, \textbf{PatchTST} \cite{nie2022time}, and \textbf{iTransformer} \cite{liu2023itransformer}. All backbones are trained using the official configurations released with the Time-Series-Library benchmark~\cite{wang2024tssurvey}, with hyperparameters taken directly from the original papers where available.

\paragraph {Training Protocol.} FSE is trained in two stages, consistent with its post-hoc design (see Proposed Method). Each baseline is first trained to convergence and frozen; FSE is then trained on top of the frozen baseline using \(\mathcal{L}_{\mathrm{total}}\). This ensures that the reported improvements are attributable to spectral correction alone, not to any additional gradient signal reaching the backbone. 

\paragraph {Experiment Details.} For all models and datasets, the prediction length was set as \(T \in\{96,192,336,720\}\), and the lookback window size \(L\) was fixed at 96. Each experiment was repeated three times for robustness. Further details about implementation and hyperparameter setting are provided in the Appendix.

\subsection{Main Results}

Table~\ref{tab:fse_all_models} reports average performance over the four prediction lengths across seven datasets and six backbones. FSE improves the majority of backbone–dataset combinations on both MSE and MAE, with the largest gains observed on TimesNet (e.g., 34.8\% MSE reduction on Traffic, 25.7\% MAE reduction on Solar), which also exhibits the most pronounced over-smoothing among the tested backbones (Figure~\ref{fig:1}). Gains are smaller but still consistent for architectures that already model periodic structure more explicitly, such as TimeMixer and Dualformer.

A small number of settings show mild degradation, e.g., TimeMixer on ETTh2 (MSE +5.4\%) and Solar (MSE +1.6\%), Dualformer on Solar (MSE +3.7\%). We interpret these cases through Proposition 2 in the Appendix: when the residual between a strong baseline and the ground truth carries little predictable structure given \((X, Y^0)\), no correction module operating on this information can systematically reduce error in that frequency band, and the risk-control operator \(\Theta\) is designed precisely to limit the resulting degradation rather than eliminate it. Standard deviations across seeds (Appendix Table~\ref{tab:confidence_interval}) show that most gains exceed run-to-run variance, while a subset of the small negative cases fall within the noise floor.
Table~\ref{tab:average_performance_single} compares FSE with two normalization-based alternatives, RevIN~\cite{kim2021reversible} and FAN~\cite{ye2024frequencyadaptivenormalizationnonstationary}, on iTransformer, TimesNet, TimeMixer, and PatchTST across the four ETT datasets. FSE achieves the lowest horizon-averaged error in 26 of 32 backbone--dataset--metric comparisons, including one tie. Table~\ref{tab:full_posthoc} reports the per-horizon results.

\begin{table}[!t]
\setlength{\abovecaptionskip}{3pt}
\setlength{\belowcaptionskip}{0pt}
    \centering
    \renewcommand{\arraystretch}{0.92}
    \setlength{\tabcolsep}{0.5pt}
    
    \resizebox{\columnwidth}{!}{
    \begin{tabular}{l c c cccc cccc}
        \toprule
        \multirow{2}{*}{\textbf{Backbone}} & \multirow{2}{*}{\textbf{Method}} & 
        \multicolumn{2}{c}{\textbf{ETTh1}} & 
        \multicolumn{2}{c}{\textbf{ETTh2}} & 
        \multicolumn{2}{c}{\textbf{ETTm1}} & 
        \multicolumn{2}{c}{\textbf{ETTm2}} \\
        
        \cmidrule(lr){3-4} \cmidrule(lr){5-6} \cmidrule(lr){7-8} \cmidrule(lr){9-10}
        
        & & MSE & MAE & MSE & MAE & MSE & MAE & MSE & MAE \\
        \midrule
        
        \multirow{4}{*}{\textbf{iTransformer}} 
        & Base    & 0.445 & 0.434 & 0.374 & 0.399 & 0.407 & 0.410 & 0.288 & 0.332 \\
        & +RevIN  & 0.435 & 0.445 & 0.358 & 0.401 & 0.384 & 0.407 & 0.281 & 0.337 \\
        & +FAN    & 0.442 & 0.446 & 0.358 & 0.404 & 0.383 & 0.409 & 0.285 & 0.332 \\
        & \textbf{+FSE (Ours)} & \textbf{0.421} & \textbf{0.432} & \textbf{0.345} & \textbf{0.388} & \textbf{0.369} & \textbf{0.397} & \textbf{0.265} & \textbf{0.310} \\
        
        \midrule
        
        \multirow{4}{*}{\textbf{TimesNet}} 
        & Base    & 0.458 & 0.450 & 0.414 & 0.427 & 0.388 & 0.406 & 0.291 & 0.333 \\
        & +RevIN  & 0.447 & 0.446 & 0.388 & 0.416 & 0.369 & 0.394 & 0.287 & 0.342 \\
        & +FAN    & 0.417 & 0.435 & 0.344 & 0.391 & 0.361 & 0.391 & 0.276 & 0.330 \\
        & \textbf{+FSE (Ours)} & \textbf{0.408} & \textbf{0.427} & \textbf{0.336} & \textbf{0.387} & \textbf{0.350} & \textbf{0.382} & \textbf{0.262} & \textbf{0.321} \\
        
\midrule
\multirow{4}{*}{\textbf{TimeMixer}} & Base & 0.411 & 0.423 & \textbf{0.316} & 0.384 & 0.348 & 0.376 & 0.256 & 0.316 \\
 & +RevIN & \textbf{0.401} & 0.426 & 0.343 & 0.385 & 0.353 & 0.383 & 0.258 & 0.316 \\
 & +FAN & 0.411 & 0.421 & 0.342 & 0.398 & 0.352 & \textbf{0.364} & 0.296 & 0.363 \\
 & \textbf{+FSE (Ours)} & \textbf{0.401} & \textbf{0.418} & 0.333 & \textbf{0.381} & \textbf{0.342} & 0.367 & \textbf{0.243} & \textbf{0.302} \\
\midrule
\multirow{4}{*}{\textbf{PatchTST}} & Base & 0.418 & 0.437 & 0.343 & 0.388 & \textbf{0.350} & \textbf{0.381} & 0.250 & 0.314 \\
 & +RevIN & 0.410 & \textbf{0.417} & 0.362 & 0.398 & 0.357 & 0.384 & 0.250 & 0.312 \\
 & +FAN & 0.431 & 0.441 & 0.368 & \textbf{0.386} & 0.359 & \textbf{0.381} & 0.252 & 0.312 \\
 & \textbf{+FSE (Ours)} & \textbf{0.409} & 0.418 & \textbf{0.340} & 0.388 & 0.351 & 0.383 & \textbf{0.242} & \textbf{0.310} \\
        \bottomrule
    \end{tabular}
    }
    \caption{Average MSE and MAE over four prediction lengths. Bold marks the best result per backbone, dataset, and metric, including ties.}
    \label{tab:average_performance_single}
\end{table}

\begin{table}[!t]
\setlength{\abovecaptionskip}{3pt}
\centering
\setlength{\tabcolsep}{2.5pt}
\renewcommand{\arraystretch}{0.86}
\resizebox{\columnwidth}{!}{%
\begin{tabular}{@{}cc cc cc cc cc@{}}
\toprule
\multicolumn{2}{c}{\textbf{Alignment}} 
& \multicolumn{2}{c}{\textbf{ETTh1}} 
& \multicolumn{2}{c}{\textbf{ETTm1}} 
& \multicolumn{2}{c}{\textbf{Electricity}} 
& \multicolumn{2}{c}{\textbf{Traffic}} \\

\cmidrule(lr){1-2} \cmidrule(lr){3-4} \cmidrule(lr){5-6} \cmidrule(lr){7-8} \cmidrule(lr){9-10}

Cosine & EMA & MSE & MAE & MSE & MAE & MSE & MAE & MSE & MAE \\
\midrule

\checkmark & $\times$ 
& 0.435 & 0.450 
& 0.357 & 0.377 
& 0.169 & 0.263 
& 0.440 & 0.300 \\

$\times$ & \checkmark 
& 0.432 & 0.447 
& 0.357 & 0.377 
& 0.167 & 0.261 
& 0.439 & 0.300 \\

\midrule
\checkmark & \checkmark 
& \textbf{0.430} & \textbf{0.445} 
& \textbf{0.355} & \textbf{0.375} 
& \textbf{0.164} & \textbf{0.260} 
& \textbf{0.435} & \textbf{0.296} \\

\bottomrule
\end{tabular}%
}
\caption{Alignment ablation with DLinear. \(\checkmark\) and \(\times\) indicate retained and removed terms.}
\label{tab: ablation}
\end{table}

\begin{table*}[t]
\centering
\renewcommand{\arraystretch}{1.15}
\setlength{\tabcolsep}{3pt}
\begin{tabular*}{\textwidth}{@{\extracolsep{\fill}}lcccc cc cc cc cc@{}}
\toprule
\multicolumn{5}{c}{\textbf{Configuration}}
& \multicolumn{2}{c}{\textbf{ETTh1}}
& \multicolumn{2}{c}{\textbf{ETTm1}}
& \multicolumn{2}{c}{\textbf{Electricity}}
& \multicolumn{2}{c}{\textbf{Traffic}} \\
\cmidrule(lr){1-5}
\cmidrule(lr){6-7}\cmidrule(lr){8-9}
\cmidrule(lr){10-11}\cmidrule(lr){12-13}
Variant & Distill & Recon. & Bound & Spectral
& MSE & MAE & MSE & MAE & MSE & MAE & MSE & MAE \\
\midrule
correction-only & $\times$ & -- & \checkmark & \checkmark
& 0.446 & 0.454 & 0.367 & 0.391 & 0.169 & 0.266 & 0.452 & 0.311 \\
w/o reconstruction & \checkmark & $\times$ & \checkmark & \checkmark
& 0.447 & 0.453 & 0.371 & 0.396 & 0.171 & 0.268 & 0.453 & 0.305 \\
w/o bound & \checkmark & \checkmark & $\times$ & \checkmark
& 0.459 & 0.463 & 0.371 & 0.393 & 0.168 & 0.270 & 0.471 & 0.313 \\
Time domain & -- & -- & -- & $\times$
& 0.477 & 0.477 & 0.397 & 0.415 & 0.170 & 0.266 & 0.462 & 0.310 \\
\midrule
Full FSE & \checkmark & \checkmark & \checkmark & \checkmark
& \textbf{0.430} & \textbf{0.445}
& \textbf{0.355} & \textbf{0.375}
& \textbf{0.164} & \textbf{0.260}
& \textbf{0.435} & \textbf{0.296} \\
\bottomrule
\end{tabular*}
\caption{Component ablations with DLinear, averaged over \(T\in\{96,192,336,720\}\). The symbols \(\checkmark\) and \(\times\) indicate retained and removed components. Dashes mark settings outside the scope of the corresponding control. A cross under Spectral denotes matched-capacity time-domain correction.}
\label{tab:average_component_ablation}
\end{table*}

\begin{table}[t]
    \centering
    
    \resizebox{\columnwidth}{!}{
    \begin{tabular}{l c cccc}
        \toprule
        \multirow{2}{*}{\textbf{Dataset}} & \multirow{2}{*}{\textbf{Pred. Len.}} & \multicolumn{2}{c}{\textbf{HF-MSE}} & \multicolumn{2}{c}{\textbf{HF-MAE}} \\
        \cmidrule(lr){3-4} \cmidrule(lr){5-6}
        
        & & \textbf{+FSE} & \textbf{Baseline} & \textbf{+FSE} & \textbf{Baseline} \\
        \midrule
        
        \multirow{4}{*}{\emph{ETTh1}} 
        & 96  & \textbf{2.17} & 2.26 & \textbf{1.04} & 1.12 \\
        & 192 & \textbf{2.57} & 2.67 & \textbf{1.18} & 1.26 \\
        & 336 & \textbf{3.00} & 3.12 & \textbf{1.32} & 1.41 \\
        & 720 & \textbf{3.69} & 3.83 & \textbf{1.54} & 1.65 \\
        \midrule 
        
        \multirow{4}{*}{\emph{ETTm1}} 
        & 96  & \textbf{1.93} & 2.07 & \textbf{0.98} & 1.11 \\
        & 192 & \textbf{2.29} & 2.46 & \textbf{1.11} & 1.26 \\
        & 336 & \textbf{2.67} & 2.87 & \textbf{1.24} & 1.40 \\
        & 720 & \textbf{3.28} & 3.52 & \textbf{1.44} & 1.63 \\
        \midrule
        
        \multirow{4}{*}{\emph{Electricity}} 
        & 96  & \textbf{30.86} & 32.49 & \textbf{4.17} & 4.46 \\
        & 192 & \textbf{36.54} & 38.48 & \textbf{4.73} & 5.06 \\
        & 336 & \textbf{42.63} & 44.89 & \textbf{5.29} & 5.66 \\
        & 720 & \textbf{52.37} & 55.15 & \textbf{6.16} & 6.58 \\
        \midrule
        
        \multirow{4}{*}{\emph{Traffic}} 
        & 96  & \textbf{192.82} & 198.69 & \textbf{8.72} & 9.48 \\
        & 192 & \textbf{228.34} & 235.29 & \textbf{9.90} & 10.75 \\
        & 336 & \textbf{266.40} & 274.50 & \textbf{11.07} & 12.02 \\
        & 720 & \textbf{327.29} & 337.24 & \textbf{12.87} & 13.99 \\
        \bottomrule
    \end{tabular}
    }
    \caption{High-frequency forecasting errors across prediction lengths. \textbf{Bold} indicates the best results.}
    \label{tab:hf_errors_by_horizon}
\end{table}

\begin{table}[t]
    \centering
    \setlength{\tabcolsep}{2.5pt}
    \renewcommand{\arraystretch}{0.90}
    \resizebox{\columnwidth}{!}{
    \begin{tabular}{@{}l c c c c@{}}
        \toprule
        \multirow{2}{*}{\textbf{Dataset}} & \multicolumn{2}{c}{\textbf{DLinear+FSE}} & \multicolumn{2}{c}{\textbf{PatchTST+FSE}} \\
        \cmidrule(lr){2-3} \cmidrule(lr){4-5} 
        
        & \textbf{Time (ms)} & \textbf{Memory (MB)} & \textbf{Time (ms)} & \textbf{Memory (MB)} \\
        \midrule
        
        \emph{ETTh1}       & +0.16 & +4.36 & +0.64 & +5.59 \\
        \emph{Electricity} & +2.09 & +11.5 & +3.35 & +19.4 \\
        
        \bottomrule
    \end{tabular}
    }
    \caption{Inference time and memory overheads introduced by FSE under 96-720 forecasting setting.}
    \label{tab:overhead}
\end{table}

\subsection{Ablation Study}

Table~\ref{tab: ablation} reports average MSE and MAE over the four prediction lengths for the DLinear ablation. Combining sample-wise cosine alignment and EMA marginal regularization yields the lowest average errors on all four datasets among the alignment variants. EMA-only alignment yields lower average errors than cosine-only alignment on ETTh1 and Electricity.

Table~\ref{tab:average_component_ablation} compares five DLinear configurations. \emph{+FSE} uses the complete method. \emph{correction-only} retains spectral residual correction and removes future distillation. \emph{w/o reconstruction} removes the teacher reconstruction loss \(\mathcal{L}_{\mathrm{recon}}\), and \emph{w/o bound} removes the relative cap on spectral corrections. \emph{Time domain} uses a matched-capacity time-domain corrector. Relative to full FSE, these four controls raise MSE by 3.51\%, 4.22\%, 5.49\%, and 8.16\%, respectively, averaging the dataset-level relative increases. Full FSE has the lowest average MSE and MAE on all four datasets. Appendix Table~\ref{tab:full_ablation} gives per-horizon results.

We also examined sensitivity to \(\eta\) (overall alignment weight) and \(\lambda_{\mathrm{cos}}\) (cosine/EMA balance ) in Appendix Figures~\ref{align_weight_sensitivity}-\ref{cosine_weight_sensitivity}. Both curves show a stable optimum for most datasets, shifting modestly with prediction horizon. For example, Longer horizons on ETTh1 benefit from a larger \(\eta\), while Electricity and Traffic prefer intermediate values. The ratio implied by the Pearson \(\chi^2\) derivation, \(\lambda_{\mathrm{cos}}\approx 0.667\), falls within the near-optimal region, though the empirical optimum is between \([0.6,0.8]\) and drifts toward 0.8 in most cases. This indicates that default settings generalize reasonably without per-dataset tuning, though further gains are available with dataset-specific search.

\subsection{High-frequency Error and Efficiency}
To check whether FSE specifically targets over-smoothing rather than improving aggregate error incidentally, we restricted evaluation in the frequency domain, focusing only on the high-frequency band identified via knee-point detection on smoothed cumulative energy~\cite{5961514, hu2026bridgingpastfuturedistributionaware}. We reported the high-frequency MSE and MAE in Table~\ref{tab:hf_errors_by_horizon}, using DLinear as the backbone. FSE reduces both metrics at all four horizons on all four datasets, with the relative gain roughly stable across horizons. 

Moreover, Table~\ref{tab:overhead} reports the inference overhead introduced by FSE under the 96-720 forecasting setting. With the teacher branch and reconstruction head dropped at inference, only the student branch and correction network add cost. As a result: on ETTh1, FSE adds only \(0.16\) ms and \(4.36\) MB to DLinear, and \(0.64\) ms and \(5.59\) MB to PatchTST. On the larger Electricity dataset, the overhead remains moderate, with at most \(3.35\) ms and \(19.4\) MB. These results support the efficiency of FSE as a lightweight post-hoc enhancement module.

\section{Conclusion}

In this paper, we proposed FSE, a lightweight, model-agnostic framework for post-hoc spectral correction in long-term forecasting. Rather than modifying the forecasting backbone, FSE refines an existing prediction directly in the frequency domain, using future-distilled guidance obtained without access to the future at inference time. A teacher branch encodes the ground-truth future spectrum during training and is distilled into a past-conditioned student representation through a non-contrastive dual-alignment objective. At inference, only the student branch remains, guiding a correction network that adjusts the baseline spectrum with a bounded, sample-dependent magnitude. Across seven benchmarks and six forecasting backbones, FSE consistently improves accuracy, reduces high-frequency error specifically, and adds negligible inference cost. These results suggest that predictable spectral residuals are an important source of improvement for long-term forecasting, and that future-distilled post-hoc correction is a practical way to recover them. Future work may extend FSE to more adaptive correction mechanisms, richer phase-aware spectral modeling, and broader sequence modeling tasks beyond long-term forecasting.

\section*{Acknowledgments}

This work was supported by JSPS KAKENHI Grant Numbers JP26H02498 and JP22H05106.

\bibliography{main}

\begin{thebibliography}{46}
\providecommand{\natexlab}[1]{#1}

\bibitem[{Alaa and van~der Schaar(2019)}]{alaa2019attentive}
Alaa, A.~M.; and van~der Schaar, M. 2019.
\newblock Attentive state-space modeling of disease progression.
\newblock \emph{Advances in neural information processing systems}, 32.

\bibitem[{Amari(2009)}]{amari2009alpha}
Amari, S.-I. 2009.
\newblock $alpha$-Divergence is unique, belonging to both $ f $-divergence and Bregman divergence classes.
\newblock \emph{IEEE Transactions on Information Theory}, 55(11): 4925--4931.

\bibitem[{Ariyo, Adewumi, and Ayo(2014)}]{ariyo2014stock}
Ariyo, A.~A.; Adewumi, A.~O.; and Ayo, C.~K. 2014.
\newblock Stock price prediction using the ARIMA model.
\newblock In \emph{2014 UKSim-AMSS 16th int'l conf. on computer modelling and simulation}, 106--112. IEEE.

\bibitem[{Bai and Kawahara(2026)}]{bai2026dualformer}
Bai, J.; and Kawahara, Y. 2026.
\newblock Dualformer: Time-Frequency Dual Domain Learning for Long-term Time Series Forecasting.
\newblock \emph{arXiv preprint arXiv:2601.15669}.

\bibitem[{Bilal et~al.(2022)Bilal, Kim, Fayaz, and Pawar}]{bilal2022comparative}
Bilal, M.; Kim, H.; Fayaz, M.; and Pawar, P. 2022.
\newblock Comparative analysis of time series forecasting approaches for household electricity consumption prediction.
\newblock \emph{arXiv preprint arXiv:2207.01019}.

\bibitem[{Chen et~al.(2020)Chen, Kornblith, Norouzi, and Hinton}]{chen2020simple}
Chen, T.; Kornblith, S.; Norouzi, M.; and Hinton, G. 2020.
\newblock A simple framework for contrastive learning of visual representations.
\newblock In \emph{International conference on machine learning}, 1597--1607. PmLR.

\bibitem[{Chen and He(2021)}]{chen2021exploring}
Chen, X.; and He, K. 2021.
\newblock Exploring simple siamese representation learning.
\newblock In \emph{Proceedings of the IEEE/CVF conference on computer vision and pattern recognition}, 15750--15758.

\bibitem[{Cichocki and Amari(2010)}]{cichocki2010families}
Cichocki, A.; and Amari, S.-i. 2010.
\newblock Families of alpha-beta-and gamma-divergences: Flexible and robust measures of similarities.
\newblock \emph{Entropy}, 12(6): 1532--1568.

\bibitem[{Das et~al.(2024)Das, Kong, Sen, and Zhou}]{das2024decoderonlyfoundationmodeltimeseries}
Das, A.; Kong, W.; Sen, R.; and Zhou, Y. 2024.
\newblock A decoder-only foundation model for time-series forecasting.
\newblock arXiv:2310.10688.

\bibitem[{Duchon and Hale(2012)}]{duchon2012time}
Duchon, C.; and Hale, R. 2012.
\newblock \emph{Time series analysis in meteorology and climatology: an introduction}.
\newblock John Wiley \& Sons.

\bibitem[{Fan et~al.(2023)Fan, Wang, Wang, Wang, Zhou, and Fu}]{fan2023dishtsgeneralparadigmalleviating}
Fan, W.; Wang, P.; Wang, D.; Wang, D.; Zhou, Y.; and Fu, Y. 2023.
\newblock Dish-TS: A General Paradigm for Alleviating Distribution Shift in Time Series Forecasting.
\newblock arXiv:2302.14829.

\bibitem[{Fatima and Rahimi(2024)}]{machines12060380}
Fatima, S. S.~W.; and Rahimi, A. 2024.
\newblock A Review of Time-Series Forecasting Algorithms for Industrial Manufacturing Systems.
\newblock \emph{Machines}, 12(6).

\bibitem[{HaoChen et~al.(2021)HaoChen, Wei, Gaidon, and Ma}]{haochen2021provable}
HaoChen, J.~Z.; Wei, C.; Gaidon, A.; and Ma, T. 2021.
\newblock Provable guarantees for self-supervised deep learning with spectral contrastive loss.
\newblock \emph{Advances in neural information processing systems}, 34: 5000--5011.

\bibitem[{Hu et~al.(2026)Hu, Yang, Zhou, Liu, Tang, Jin, and Sun}]{hu2026bridgingpastfuturedistributionaware}
Hu, Y.; Yang, J.; Zhou, T.; Liu, P.; Tang, Y.; Jin, R.; and Sun, L. 2026.
\newblock Bridging Past and Future: Distribution-Aware Alignment for Time Series Forecasting.
\newblock arXiv:2509.14181.

\bibitem[{Kim et~al.(2021)Kim, Kim, Tae, Park, Choi, and Choo}]{kim2021reversible}
Kim, T.; Kim, J.; Tae, Y.; Park, C.; Choi, J.-H.; and Choo, J. 2021.
\newblock Reversible instance normalization for accurate time-series forecasting against distribution shift.
\newblock In \emph{Int'l Conf. on Learning Representations}.

\bibitem[{Lai et~al.(2018)Lai, Chang, Yang, and Liu}]{lai2018modelinglongshorttermtemporal}
Lai, G.; Chang, W.-C.; Yang, Y.; and Liu, H. 2018.
\newblock Modeling Long- and Short-Term Temporal Patterns with Deep Neural Networks.
\newblock arXiv:1703.07015.

\bibitem[{Li et~al.(2024)Li, Qin, Cheng, and Tan}]{li2024ftmixer}
Li, Z.; Qin, Y.; Cheng, X.; and Tan, Y. 2024.
\newblock Ftmixer: Frequency and time domain representations fusion for time series modeling.
\newblock \emph{arXiv preprint arXiv:2405.15256}.

\bibitem[{Liu et~al.(2023{\natexlab{a}})Liu, Hu, Zhang, Wu, Wang, Ma, and Long}]{liu2023itransformer}
Liu, Y.; Hu, T.; Zhang, H.; Wu, H.; Wang, S.; Ma, L.; and Long, M. 2023{\natexlab{a}}.
\newblock itransformer: Inverted transformers are effective for time series forecasting.
\newblock \emph{arXiv preprint arXiv:2310.06625}.

\bibitem[{Liu et~al.(2023{\natexlab{b}})Liu, Li, Wang, and Long}]{liu2023koopa}
Liu, Y.; Li, C.; Wang, J.; and Long, M. 2023{\natexlab{b}}.
\newblock Koopa: Learning non-stationary time series dynamics with koopman predictors.
\newblock \emph{Advances in neural information processing systems}, 36: 12271--12290.

\bibitem[{Liu et~al.(2023{\natexlab{c}})Liu, Wu, Wang, and Long}]{liu2023nonstationarytransformersexploringstationarity}
Liu, Y.; Wu, H.; Wang, J.; and Long, M. 2023{\natexlab{c}}.
\newblock Non-stationary Transformers: Exploring the Stationarity in Time Series Forecasting.
\newblock arXiv:2205.14415.

\bibitem[{Nachum and Dai(2020)}]{nachum2020reinforcement}
Nachum, O.; and Dai, B. 2020.
\newblock Reinforcement learning via fenchel-rockafellar duality.
\newblock \emph{arXiv preprint arXiv:2001.01866}.

\bibitem[{Nguyen, Wainwright, and Jordan(2010)}]{nguyen2010estimating}
Nguyen, X.; Wainwright, M.~J.; and Jordan, M.~I. 2010.
\newblock Estimating divergence functionals and the likelihood ratio by convex risk minimization.
\newblock \emph{IEEE Transactions on Information Theory}, 56(11): 5847--5861.

\bibitem[{Nie et~al.(2022)Nie, Nguyen, Sinthong, and Kalagnanam}]{nie2022time}
Nie, Y.; Nguyen, N.~H.; Sinthong, P.; and Kalagnanam, J. 2022.
\newblock A time series is worth 64 words: Long-term forecasting with transformers.
\newblock \emph{arXiv preprint arXiv:2211.14730}.

\bibitem[{Oord, Li, and Vinyals(2018)}]{oord2018representation}
Oord, A. v.~d.; Li, Y.; and Vinyals, O. 2018.
\newblock Representation learning with contrastive predictive coding.
\newblock \emph{arXiv preprint arXiv:1807.03748}.

\bibitem[{Passalis et~al.(2019)Passalis, Tefas, Kanniainen, Gabbouj, and Iosifidis}]{passalis2019deepadaptiveinputnormalization}
Passalis, N.; Tefas, A.; Kanniainen, J.; Gabbouj, M.; and Iosifidis, A. 2019.
\newblock Deep Adaptive Input Normalization for Time Series Forecasting.
\newblock arXiv:1902.07892.

\bibitem[{Piao et~al.(2024)Piao, Chen, Dong, Matsubara, and Sakurai}]{piao2024frednormerfrequencydomainnormalization}
Piao, X.; Chen, Z.; Dong, Y.; Matsubara, Y.; and Sakurai, Y. 2024.
\newblock FredNormer: Frequency Domain Normalization for Non-stationary Time Series Forecasting.
\newblock arXiv:2410.01860.

\bibitem[{Satopaa et~al.(2011)Satopaa, Albrecht, Irwin, and Raghavan}]{5961514}
Satopaa, V.; Albrecht, J.; Irwin, D.; and Raghavan, B. 2011.
\newblock Finding a "Kneedle" in a Haystack: Detecting Knee Points in System Behavior.
\newblock In \emph{2011 31st International Conference on Distributed Computing Systems Workshops}, 166--171.

\bibitem[{Urettini et~al.(2024)Urettini, Atzeni, Ramjattan, and Carta}]{Urettini_2024}
Urettini, E.; Atzeni, D.; Ramjattan, R.~J.; and Carta, A. 2024.
\newblock GAS-Norm: Score-Driven Adaptive Normalization for Non-Stationary Time Series Forecasting in Deep Learning.
\newblock In \emph{Proceedings of the 33rd ACM International Conference on Information and Knowledge Management}, CIKM ’24, 2282–2291. ACM.

\bibitem[{Wang et~al.(2024{\natexlab{a}})Wang, Wu, Shi, Hu, Luo, Ma, Zhang, and Zhou}]{wang2024timemixerdecomposablemultiscalemixing}
Wang, S.; Wu, H.; Shi, X.; Hu, T.; Luo, H.; Ma, L.; Zhang, J.~Y.; and Zhou, J. 2024{\natexlab{a}}.
\newblock TimeMixer: Decomposable Multiscale Mixing for Time Series Forecasting.
\newblock arXiv:2405.14616.

\bibitem[{Wang et~al.(2024{\natexlab{b}})Wang, Wu, Dong, Liu, Long, and Wang}]{wang2024tssurvey}
Wang, Y.; Wu, H.; Dong, J.; Liu, Y.; Long, M.; and Wang, J. 2024{\natexlab{b}}.
\newblock Deep Time Series Models: A Comprehensive Survey and Benchmark.

\bibitem[{Woo et~al.(2024)Woo, Liu, Kumar, Xiong, Savarese, and Sahoo}]{woo2024moirai}
Woo, G.; Liu, C.; Kumar, A.; Xiong, C.; Savarese, S.; and Sahoo, D. 2024.
\newblock Unified Training of Universal Time Series Forecasting Transformers.
\newblock In \emph{Forty-first International Conference on Machine Learning}.

\bibitem[{Woo et~al.(2022)Woo, Liu, Sahoo, Kumar, and Hoi}]{woo2022costcontrastivelearningdisentangled}
Woo, G.; Liu, C.; Sahoo, D.; Kumar, A.; and Hoi, S. 2022.
\newblock CoST: Contrastive Learning of Disentangled Seasonal-Trend Representations for Time Series Forecasting.
\newblock arXiv:2202.01575.

\bibitem[{Wu et~al.(2022)Wu, Hu, Liu, Zhou, Wang, and Long}]{wu2022timesnet}
Wu, H.; Hu, T.; Liu, Y.; Zhou, H.; Wang, J.; and Long, M. 2022.
\newblock Timesnet: Temporal 2d-variation modeling for general time series analysis.
\newblock \emph{arXiv preprint arXiv:2210.02186}.

\bibitem[{Wu et~al.(2021)Wu, Xu, Wang, and Long}]{wu2021autoformer}
Wu, H.; Xu, J.; Wang, J.; and Long, M. 2021.
\newblock Autoformer: Decomposition transformers with auto-correlation for long-term series forecasting.
\newblock \emph{Advances in neural information processing systems}, 34: 22419--22430.

\bibitem[{Xu, Zeng, and Xu(2023)}]{xu2023fits}
Xu, Z.; Zeng, A.; and Xu, Q. 2023.
\newblock FITS: Modeling time series with $10 k $ parameters.
\newblock \emph{arXiv preprint arXiv:2307.03756}.

\bibitem[{Ye et~al.(2024)Ye, Deng, Zou, and Gui}]{ye2024frequencyadaptivenormalizationnonstationary}
Ye, W.; Deng, S.; Zou, Q.; and Gui, N. 2024.
\newblock Frequency Adaptive Normalization For Non-stationary Time Series Forecasting.
\newblock arXiv:2409.20371.

\bibitem[{Yi et~al.(2024)Yi, Zhang, Fan, Wang, Wang, He, An, Lian, Cao, and Niu}]{yi2024frequency}
Yi, K.; Zhang, Q.; Fan, W.; Wang, S.; Wang, P.; He, H.; An, N.; Lian, D.; Cao, L.; and Niu, Z. 2024.
\newblock Frequency-domain MLPs are more effective learners in time series forecasting.
\newblock \emph{Advances in Neural Information Processing Systems}, 36.

\bibitem[{Yue et~al.(2025)Yue, Liu, Ying, Xing, Guo, and Shi}]{yue2025freeformer}
Yue, W.; Liu, Y.; Ying, X.; Xing, B.; Guo, R.; and Shi, J. 2025.
\newblock Freeformer: Frequency enhanced transformer for multivariate time series forecasting.
\newblock \emph{arXiv preprint arXiv:2501.13989}.

\bibitem[{Yue et~al.(2022)Yue, Wang, Duan, Yang, Huang, Tong, and Xu}]{yue2022ts2vecuniversalrepresentationtime}
Yue, Z.; Wang, Y.; Duan, J.; Yang, T.; Huang, C.; Tong, Y.; and Xu, B. 2022.
\newblock TS2Vec: Towards Universal Representation of Time Series.
\newblock arXiv:2106.10466.

\bibitem[{Zeng et~al.(2023)Zeng, Chen, Zhang, and Xu}]{zeng2023transformers}
Zeng, A.; Chen, M.; Zhang, L.; and Xu, Q. 2023.
\newblock Are transformers effective for time series forecasting?
\newblock In \emph{Proceedings of the AAAI conf. on artificial intelligence}, volume~37, 11121--11128.

\bibitem[{Zhang and Xiao(2026)}]{zhang2026norinbackboneadaptivereversiblenormalization}
Zhang, S.; and Xiao, Y. 2026.
\newblock NoRIN: Backbone-Adaptive Reversible Normalization for Time-Series Forecasting.
\newblock arXiv:2605.10823.

\bibitem[{Zhang et~al.(2022)Zhang, Zhao, Tsiligkaridis, and Zitnik}]{zhang2022selfsupervisedcontrastivepretrainingtime}
Zhang, X.; Zhao, Z.; Tsiligkaridis, T.; and Zitnik, M. 2022.
\newblock Self-Supervised Contrastive Pre-Training For Time Series via Time-Frequency Consistency.
\newblock arXiv:2206.08496.

\bibitem[{Zheng et~al.(2024)Zheng, Chen, Schürch, Mollaysa, Allam, and Krauthammer}]{zheng2024simplecontrastiverepresentationlearning}
Zheng, X.; Chen, X.; Schürch, M.; Mollaysa, A.; Allam, A.; and Krauthammer, M. 2024.
\newblock Simple Contrastive Representation Learning for Time Series Forecasting.
\newblock arXiv:2303.18205.

\bibitem[{Zhou et~al.(2021)Zhou, Zhang, Peng, Zhang, Li, Xiong, and Zhang}]{zhou2021informer}
Zhou, H.; Zhang, S.; Peng, J.; Zhang, S.; Li, J.; Xiong, H.; and Zhang, W. 2021.
\newblock Informer: Beyond efficient transformer for long sequence time-series forecasting.
\newblock In \emph{Proceedings of the AAAI conf. on artificial intelligence}, volume~35, 11106--11115.

\bibitem[{Zhou et~al.(2022{\natexlab{a}})Zhou, Ma, Wen, Sun, Yao, Yin, Jin et~al.}]{zhou2022film}
Zhou, T.; Ma, Z.; Wen, Q.; Sun, L.; Yao, T.; Yin, W.; Jin, R.; et~al. 2022{\natexlab{a}}.
\newblock Film: Frequency improved legendre memory model for long-term time series forecasting.
\newblock \emph{Advances in neural information processing systems}, 35: 12677--12690.

\bibitem[{Zhou et~al.(2022{\natexlab{b}})Zhou, Ma, Wen, Wang, Sun, and Jin}]{zhou2022fedformer}
Zhou, T.; Ma, Z.; Wen, Q.; Wang, X.; Sun, L.; and Jin, R. 2022{\natexlab{b}}.
\newblock Fedformer: Frequency enhanced decomposed transformer for long-term series forecasting.
\newblock In \emph{Int'l conf. on machine learning}, 27268--27286. PMLR.

\end{thebibliography}

\clearpage

\appendix

\section*{Appendix}

\section{Generalization of Dual Alignment Objective}

We further show that the proposed dual-alignment objective is not restricted to the Pearson \(\chi^2\) divergence. Instead, it can be viewed as one instance of a family of divergence-induced objectives whose marginal term admits a unified EMA-based non-contrastive form.

Following the previously introduced \(f\)-MI formulation, we parameterize the variational score by the similarity between the student and teacher representations:
\[
u(x,y)=z_x^\top z_y,
\qquad
\|z_x\|_2=\|z_y\|_2=1.
\]
The corresponding objective can be written as
\[
\max_{\phi}
\mathbb{E}_{p(x,y)}
\left[
t(u(x,y))
\right]
-
\mathbb{E}_{p(x)p(y)}
\left[
f^*(t(u(x,y)))
\right].
\]
The first term aligns matched past-future pairs, while the second term involves samples drawn from the product of marginals. In mini-batch training, this marginal term usually requires explicit cross-sample pairwise comparison, which leads to a contrastive objective.

We consider the \(\alpha\)-divergence family \cite{amari2009alpha, cichocki2010families}:
\[
f_\alpha(r)
=
\frac{r^\alpha-\alpha r+\alpha-1}{\alpha(\alpha-1)},
\qquad
\alpha\in\mathbb{R}/\{0,1\}.
\]
And its Fenchel conjugate
\[
f_\alpha^{*}(v)
=
\frac{1}{\alpha}
\left|1+(\alpha-1)v\right|^{\frac{\alpha}{\alpha-1}}
-
\frac{1}{\alpha}.
\]
This family includes several common divergences as special or limiting cases. In particular, \(\alpha=2\) recovers the Pearson \(\chi^2\) divergence up to a positive scaling factor, while the limits \(\alpha\rightarrow 1\) and \(\alpha\rightarrow 0\) correspond to KL and reverse KL divergences, respectively.

For this family, we use the following score transformation:
\[
t_\alpha(u)
=
\frac{\operatorname{sgn}(u)}{\alpha-1}
\left|
\sqrt{\frac{\alpha}{2}}u
\right|^{\frac{2(\alpha-1)}{\alpha}}
-
\frac{1}{\alpha-1}.
\]
With the Fenchel conjugate of the signed \(\alpha\)-divergence extension, this transformation gives
\[
f_\alpha^*(t_\alpha(u))
=
\frac{1}{2}u^2
-
\frac{1}{\alpha}.
\]
Therefore, the corresponding objective becomes, up to constants,
\[
\max_{\phi}
\mathbb{E}_{p(x,y)}
\left[
t_\alpha(z_x^\top z_y)
\right]
-
\frac{1}{2}
\mathbb{E}_{p(x)p(y)}
\left[
(z_x^\top z_y)^2
\right].
\]

The second term is the source of explicit contrastive pair construction. However, it can be rewritten as
\begin{align*}
\mathbb{E}_{p(x)p(y)}
\left[
(z_x^\top z_y)^2
\right]
&=
\mathbb{E}_{p(x)p(y)}
\left[
z_x^\top z_y z_y^\top z_x
\right] \notag \\
&=
\mathbb{E}_{p(x)}
\left[
z_x^\top
\Lambda_y
z_x
\right],
\end{align*}
where
\[
\Lambda_y
=
\mathbb{E}_{p(y)}
\left[
z_yz_y^\top
\right].
\]
Thus, the product-of-marginals term can be estimated without explicit negative pairs by maintaining an EMA teacher second-moment matrix:
\[
\Lambda_y^{(t)}
=
\beta\Lambda_y^{(t-1)}
+
(1-\beta)
\frac{1}{B}
\sum_{i=1}^{B}
z_{y,i}^{(t)}z_{y,i}^{(t)\top}.
\]

This leads to the generalized non-contrastive dual-alignment loss:
\begin{equation}
\mathcal{L}_{\mathrm{align}}^{(\alpha)}
=
-
\frac{1}{B}
\sum_{i=1}^{B}
t_\alpha
\left(
z_{x,i}^{\top}\operatorname{sg}(z_{y,i})
\right)
+
\frac{1}{2B}
\sum_{i=1}^{B}
z_{x,i}^{\top}
\operatorname{sg}(\Lambda_y)
z_{x,i}.
\label{eq:generalized_da}
\end{equation}
Different choices of \(\alpha\) induce different matched-pair alignment geometries through \(t_\alpha(\cdot)\), while the marginal regularization keeps the same EMA quadratic form. It shows that the dual-alignment objective is not tied to the Pearson \(\chi^2\) case, and the explicit contrastive marginal term can be consistently converted into a non-contrastive EMA regularizer.

\section{Additional Theoretical Analysis}
\label{app:theory}

Here we provide two supporting analyses for FSE. The first connects the student-teacher latent gap with the spectral correction error. The second clarifies that the achievable improvement depends on the predictable component of the spectral residual.

\subsection{Latent Alignment and Correction Error}
\label{app:latent_alignment_bound}

Let \(Y\) be the ground-truth future sequence and \(\hat{Y}^{0}\) be the baseline forecast. Their spectra are denoted by
\[
S_y=\mathcal{F}(Y),
\qquad
\hat{S}^{0}=\mathcal{F}(\hat{Y}^{0}).
\]
The ideal spectral residual is
\[
\Delta S^{*}=S_y-\hat{S}^{0}.
\]
FSE predicts a spectral correction using the student latent \(z_x\):
\[
\widetilde{S}
=
\hat{S}^{0}
+
A_{\theta}(z_x,\hat{S}^{0}),
\]
where \(A_{\theta}\) denotes the spectral correction network. The corrected prediction is
\[
\widetilde{Y}
=
\mathcal{F}^{-1}(\widetilde{S}).
\]

\noindent\textbf{Proposition 1.}
Assume that, for a fixed baseline spectrum \(\hat{S}^{0}\), the correction network \(A_{\theta}(z,\hat{S}^{0})\) is \(L_A\)-Lipschitz with respect to its latent input:
\[
\left\|
A_{\theta}(z_1,\hat{S}^{0})
-
A_{\theta}(z_2,\hat{S}^{0})
\right\|_2
\leq
L_A
\left\|
z_1-z_2
\right\|_2.
\]
Then the correction error satisfies
\begin{align*}
\left\|
Y-\widetilde{Y}
\right\|_2
& \leq \\
& c_{\mathcal{F}}
\left(
\left\|
\Delta S^{*}
-
A_{\theta}(z_y,\hat{S}^{0})
\right\|_2
+
L_A
\left\|
z_y-z_x
\right\|_2
\right),
\end{align*}

where \(z_y\) is the future teacher latent and \(c_{\mathcal{F}}\) is the normalization constant induced by the Fourier transform. In particular, if a unitary Fourier transform is used, then \(c_{\mathcal{F}}=1\). If the unnormalized DFT convention is used, then \(c_{\mathcal{F}}=1/\sqrt{T}\).

If \(z_x\) and \(z_y\) are \(\ell_2\)-normalized, then
\[
\left\|
z_y-z_x
\right\|_2
=
\sqrt{
2
\left(
1-z_x^\top z_y
\right)
}.
\]

\noindent\textit{Proof.}
By definition,
\begin{align*}
S_y-\widetilde{S}
& =
S_y-\hat{S}^{0}-A_{\theta}(z_x,\hat{S}^{0})\\
& =
\Delta S^{*}-A_{\theta}(z_x,\hat{S}^{0}).  
\end{align*}

Adding and subtracting \(A_{\theta}(z_y,\hat{S}^{0})\), we obtain
\begin{align*}
&\left\|S_y-\widetilde{S}\right\|_2
=
\left\|\Delta S^{*}-A_{\theta}(z_x,\hat{S}^{0})\right\|_2  \\
&=
\left\|\Delta S^{*}
-
A_{\theta}(z_y,\hat{S}^{0})
+
A_{\theta}(z_y,\hat{S}^{0})
-
A_{\theta}(z_x,\hat{S}^{0})
\right\|_2  \\
&\leq\left\|\Delta S^{*}
-
A_{\theta}(z_y,\hat{S}^{0})\right\|_2
+
\left\|A_{\theta}(z_y,\hat{S}^{0})
-
A_{\theta}(z_x,\hat{S}^{0})\right\|_2  \\
&\leq
\left\|
\Delta S^{*}
-
A_{\theta}(z_y,\hat{S}^{0})
\right\|_2
+
L_A
\left\|
z_y-z_x
\right\|_2 .
\end{align*}
By Parseval's theorem,
\[
\left\|
Y-\widetilde{Y}
\right\|_2
\leq
c_{\mathcal{F}}
\left\|
S_y-\widetilde{S}
\right\|_2 .
\]
This proves the first claim. If \(z_x\) and \(z_y\) are normalized, then
\[
\left\|
z_y-z_x
\right\|_2^2
=
\left\|
z_y
\right\|_2^2
+
\left\|
z_x
\right\|_2^2
-
2z_x^\top z_y
=
2(1-z_x^\top z_y).
\]
Taking the square root gives the result.
\(\square\)

Proposition~1 shows that the correction error is controlled by two factors: the ability of the teacher latent to generate a useful spectral residual, and the discrepancy between the student and teacher latents. Therefore, reducing the alignment gap helps tighten the correction error bound when the teacher latent contains correction-relevant spectral information.

\subsection{Predictable Spectral Residual}
\label{app:predictable_residual}

We next clarify what part of the spectral residual can be recovered by a post-hoc correction module. Let \(P_b(\cdot)\) denote the projection onto frequency band \(b\). The band-wise ideal residual is
\[
\Delta S_b^{*}
=
P_b(S_y-\hat{S}^{0}).
\]
Since FSE only uses information available at inference time, the predictable component of the residual is defined as
\[
m_b(X,\hat{S}^{0})
=
\mathbb{E}
\left[
\Delta S_b^{*}
\mid
X,\hat{S}^{0}
\right].
\]

\noindent\textbf{Proposition 2.}
For any measurable correction function \(h_b(X,\hat{S}^{0})\), we have
\begin{align*}
\mathbb{E}
\left\|
\Delta S_b^{*}
-
h_b(X,\hat{S}^{0})
\right\|_2^2
&=
\mathbb{E}
\left\|
\Delta S_b^{*}
-
m_b(X,\hat{S}^{0})
\right\|_2^2\\
&\quad+
\mathbb{E}
\left\|
m_b(X,\hat{S}^{0})
-
h_b(X,\hat{S}^{0})
\right\|_2^2.   
\end{align*}
Therefore, the optimal correction under the available information is
\[
h_b^{*}(X,\hat{S}^{0})
=
m_b(X,\hat{S}^{0}).
\]
Moreover, the maximum achievable error reduction in band \(b\) is
\[
\mathbb{E}
\left\|
\Delta S_b^{*}
\right\|_2^2
-
\mathbb{E}
\left\|
\Delta S_b^{*}
-
m_b(X,\hat{S}^{0})
\right\|_2^2
=
\mathbb{E}
\left\|
m_b(X,\hat{S}^{0})
\right\|_2^2.
\]

\noindent\textit{Proof.}
Define the unpredictable residual as
\[
\epsilon_b
=
\Delta S_b^{*}
-
m_b(X,\hat{S}^{0}).
\]
By the definition of conditional expectation,
\[
\mathbb{E}
[
\epsilon_b
\mid
X,\hat{S}^{0}
]
=
0.
\]
For any measurable \(h_b(X,\hat{S}^{0})\),
\[
\Delta S_b^{*}
-
h_b(X,\hat{S}^{0})
=
\epsilon_b
+
m_b(X,\hat{S}^{0})
-
h_b(X,\hat{S}^{0}).
\]
Thus,
\[
\begin{aligned}
&\mathbb{E}
\left\|
\Delta S_b^{*}
-
h_b(X,\hat{S}^{0})
\right\|_2^2  \\
&=
\mathbb{E}
\left\|
\epsilon_b
+
m_b(X,\hat{S}^{0})
-
h_b(X,\hat{S}^{0})
\right\|_2^2  \\
&=
\mathbb{E}
\left\|
\epsilon_b
\right\|_2^2
+
\mathbb{E}
\left\|
m_b(X,\hat{S}^{0})
-
h_b(X,\hat{S}^{0})
\right\|_2^2  \\
&\quad+
2\mathbb{E}
\left\langle
\epsilon_b,
m_b(X,\hat{S}^{0})
-
h_b(X,\hat{S}^{0})
\right\rangle .
\end{aligned}
\]
Since \(m_b(X,\hat{S}^{0})-h_b(X,\hat{S}^{0})\) is measurable with respect to \((X,\hat{S}^{0})\), the cross term is
\[
\begin{aligned}
&\mathbb{E}
\left\langle
\epsilon_b,
m_b(X,\hat{S}^{0})
-
h_b(X,\hat{S}^{0})
\right\rangle  \\
&=
\mathbb{E}
\left[
\left\langle
\mathbb{E}
[
\epsilon_b
\mid
X,\hat{S}^{0}
],
m_b(X,\hat{S}^{0})
-
h_b(X,\hat{S}^{0})
\right\rangle
\right]
=
0.
\end{aligned}
\]
Therefore,
\begin{align*}
&\mathbb{E}
\left\|
\Delta S_b^{*}
-
h_b(X,\hat{S}^{0})
\right\|_2^2\\
&=
\mathbb{E}
\left\|
\epsilon_b
\right\|_2^2
+
\mathbb{E}
\left\|
m_b(X,\hat{S}^{0})
-
h_b(X,\hat{S}^{0})
\right\|_2^2.    
\end{align*}
Since
\[
\mathbb{E}
\left\|
\Delta S_b^{*}
-
m_b(X,\hat{S}^{0})
\right\|_2^2
=
\mathbb{E}
\left\|
\epsilon_b
\right\|_2^2,
\]
the first claim follows.

Setting \(h_b(X,\hat{S}^{0})=0\), we obtain
\[
\Delta S_b^{*}
=
m_b(X,\hat{S}^{0})
+
\epsilon_b.
\]
Using the same orthogonality argument,
\[
\mathbb{E}
\left\|
\Delta S_b^{*}
\right\|_2^2
=
\mathbb{E}
\left\|
m_b(X,\hat{S}^{0})
\right\|_2^2
+
\mathbb{E}
\left\|
\epsilon_b
\right\|_2^2.
\]
Thus,
\[
\mathbb{E}
\left\|
\Delta S_b^{*}
\right\|_2^2
-
\mathbb{E}
\left\|
\Delta S_b^{*}
-
m_b(X,\hat{S}^{0})
\right\|_2^2
=
\mathbb{E}
\left\|
m_b(X,\hat{S}^{0})
\right\|_2^2.
\]
This completes the proof.
\(\square\)

Proposition~2 shows that the benefit of spectral correction is determined by the energy of the predictable residual. If a frequency band contains only unpredictable residual components, no correction module using only \(X\) and \(\hat{S}^{0}\) can systematically improve that band. This supports the design principle of FSE: it aims to recover predictable spectral residuals rather than blindly amplifying all high-frequency components.

\begin{table*}[t]
    \centering
    \begin{tabular}{c|c|c|c|c|c}
    \toprule
    Dataset & Source & Frequency & Variate & Length & Time Range  \\
    \midrule
    ETTh1 & Informer\citep{zhou2021informer} & 1h & 7 & 17420 & 2016-2017\\
    \midrule
    ETTh2 & Informer & 1h & 7 & 17420 & 2017-2018\\
    \midrule
    ETTm1 & Informer & 15min & 7 & 69680 & 2016-2017\\
    \midrule
    ETTm2 & Informer & 15min & 7 & 69680 & 2017-2018\\
    \midrule
    Electricity & UCI ML Repository & 1h & 321 & 26304 & 2012-2014\\
    \midrule
    Solar energy & \citep{lai2018modelinglongshorttermtemporal}& 10min & 137 & 52560 & 2006\\
    \midrule
    Traffic & Informer & 1h & 862 & 17544 & 2015-2016\\
    \bottomrule
    \end{tabular}
    \caption{Dataset description.}\label{tab:data}
\end{table*}

\section{Experimental Details}

\subsection{Datasets} Details about the datasets used in this work are given below. Further descriptions are in Table \ref{tab:data}. 

\noindent\textbf{ETT (Electricity Transformer Temperature)}: consists of two hourly datasets (ETTh) and two 15-minute datasets (ETTm), containing loads and oil temperature information of the electricity transformers collected from July 2016 to July 2018.

\noindent\textbf{Electricity}: electricity consumption of 321 customers recorded hourly from 2012 to 2014.

\noindent\textbf{Traffic}: road occupancy rates recorded hourly from highway sensors in San Francisco between 2015 and 2016.

\noindent\textbf{Solar Energy}: solar power production records sampled every 10 minutes from 137 PV plants in Alabama State in 2006.

\subsection{Implementation Details}
All baseline models were implemented based on the Time-Series-Library repository~\citep {wang2024tssurvey}. The experiments were conducted using PyTorch and trained on one Tesla V100 32GB GPU. \\
\noindent\textbf{Optimizer}: We employed ADAM with an initial learning rate of $1e^{-4}$, which was adjusted using a cosine decay schedule.\\
\noindent\textbf{Batchsize}: For the Traffic and Electricity datasets, batchsize was set to 32, while for all other datasets, batchsize was set to 256 to ensure efficient training while maintaining stable gradient updates.\\
\noindent\textbf{Early Stopping}: A three-epoch early stopping criterion was used to prevent overfitting.\\
\noindent\textbf{Hyperparameter Setting}: All FSE modules used \(N=3\) encoder layers, correction magnitude cap \(\alpha=0.5\), and \(\epsilon=10^{-6}\). FSE-specific loss weights are $\gamma=1$ and $\eta=0.6$,  with the dual-alignment objective set to $\lambda_{\mathrm{cos}}=0.8, \lambda_{\mathbf{ema}}=0.2$. These values were selected via validation-set sweeps reported in Appendix Figures \ref{align_weight_sensitivity}–\ref{cosine_weight_sensitivity} and used identically across all datasets and backbones. The detailed sensitivity check results and discussions are given in the following section.

\begin{figure*}
    \centering
    \includegraphics[width=\linewidth]{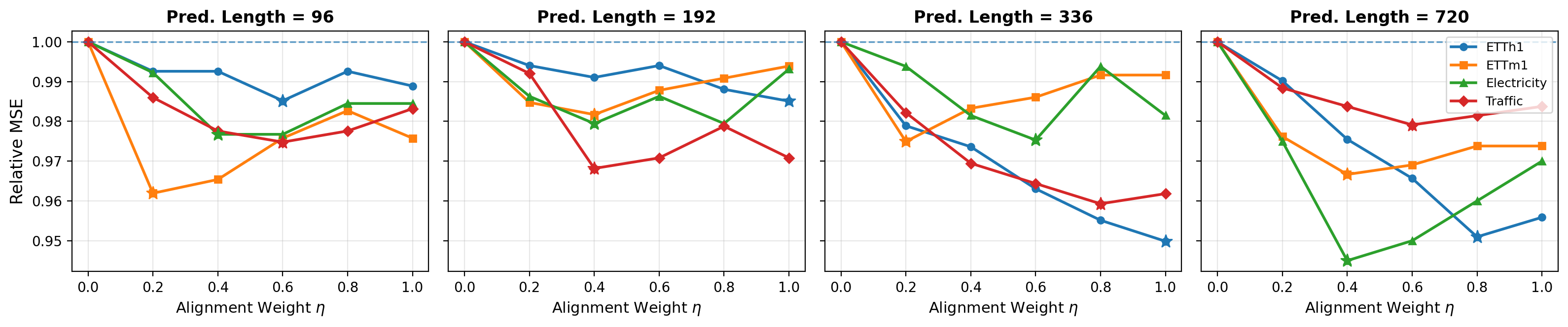}
    \caption{Sensitivity analysis of the alignment loss weight $\eta$. Lower values indicate better performance, and stars mark the best setting for each dataset-horizon curve. (Baseline: DLinear)}
    \label{align_weight_sensitivity}
\end{figure*}

\begin{figure*}
    \centering
    \includegraphics[width=\linewidth]{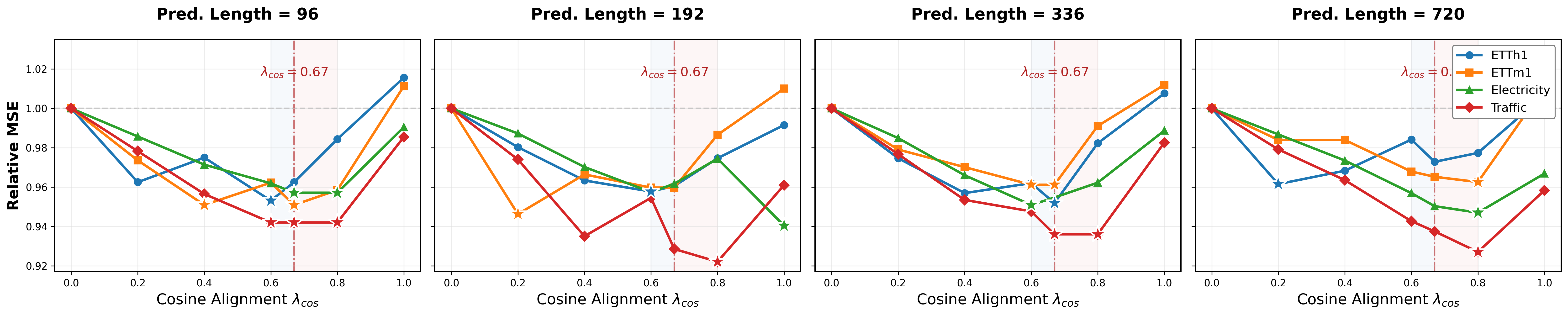}
    \caption{Sensitivity analysis of the cosine alignment weight $\lambda_{\cos}$, with $\lambda_{\mathrm{ema}}=1-\lambda_{\cos}$. Lower values indicate better performance, and stars mark the best setting for each dataset-horizon curve.  (Baseline: DLinear)}
    \label{cosine_weight_sensitivity}
\end{figure*}

\section{Hyperparameter Sensitivity}

We analyze two key hyperparameters of FSE: the overall weight of the dual-alignment loss \(\eta\), and the internal trade-off between cosine alignment and EMA alignment, controlled by \(\lambda_{\cos}\) with \(\lambda_{\mathrm{ema}}=1-\lambda_{\cos}\). For a hyperparameter \(\beta\), we report the relative MSE:
\[
\mathrm{Relative\ MSE}
=
\frac{\mathrm{MSE}(\beta)}
{\mathrm{MSE}(\beta=0)}.
\]
Thus, a value smaller than \(1\) means the corresponding loss term reduces MSE and improves forecasting performance. In the sensitivity curves, lower values indicate better performance, and the best parameter setting for each curve is marked with a star.

Figure~\ref{align_weight_sensitivity} shows that introducing the dual-alignment loss generally improves performance across datasets and prediction lengths. The optimal \(\eta\) varies across settings: shorter horizons often prefer moderate alignment strength, while longer horizons can benefit from stronger alignment on some datasets. For example, ETTh1 obtains larger gains at long horizons when \(\eta\) is high, whereas Electricity and Traffic show their best performance at intermediate values. This suggests that the alignment loss should be strong enough to transfer future spectral knowledge, but overly large weights may over-regularize the correction module for certain datasets.

Figure~\ref{cosine_weight_sensitivity} further studies the balance between local cosine alignment and global EMA alignment. Most settings achieve better results with a mixture of the two terms rather than using either extreme alone. The optimal \(\lambda_{\cos}\) is dataset- and horizon-dependent: Traffic and Electricity often favor a larger cosine weight at longer horizons, while ETT datasets show more stable behavior under moderate weights. To directly examine the ratio implied by the Pearson \(\chi^2\) derivation (\(\lambda_{\mathrm{cos}}/\lambda_{\mathrm{ema}}=2, \text{i.e., } \lambda_{\mathrm{cos}}=2/3\)), we additionally evaluate this value alongside the coarser grid. As shown in Figure~\ref{cosine_weight_sensitivity}, it falls within the flat, near-optimal region shared across datasets and prediction horizons, closely tracking the empirical optima for most cases. This indicates the theoretical ratio is a reasonable default within the empirically favorable range, rather than a value the data single out as uniquely optimal.

Overall, relative MSE across the full sweep varies only within roughly 0.92–1.02, indicating that performance is not critically sensitive to the exact weighting. The two alignment terms are complementary, with the cosine term providing matched past-future distillation and the EMA term preserving the global structure of future spectral representations. Based on the overall validation performance, we use \(\eta=0.6\) and \(\lambda_{\cos}=0.8\) by default.

\section{Model Robustness}
Table~\ref{tab:confidence_interval} reports the mean and standard deviation across three random seeds for all six backbones on all seven datasets. 

\section{Full Results}
Tables~\ref{tab:full_posthoc} and~\ref{tab:full_main_results} report the complete post-hoc comparison and the full baseline-versus-FSE results, respectively.

\subsection{Extended Component Ablation}

Table~\ref{tab:full_ablation} compares full FSE with DLinear and six ablated variants. The correction-only variant removes future distillation and retains spectral residual correction. The variants without reconstruction or the relative bound remove \(\mathcal{L}_{\mathrm{recon}}\) or the correction cap, respectively. The time-domain control uses a matched-capacity corrector.

Relative to full FSE, correction-only, removal of reconstruction, removal of the bound, and time-domain correction raise MSE by 3.51\%, 4.22\%, 5.49\%, and 8.16\%, respectively. Each percentage is the mean of the four dataset-level relative increases, computed from the horizon-averaged MSE. Full FSE also attains lower average MAE on every dataset than each of these four controls. The correction-only comparison supports the contribution of future guidance beyond spectral residual correction. The reconstruction and bound comparisons support the use of teacher reconstruction and correction magnitude control, respectively. The matched-capacity comparison favors spectral correction in the tested DLinear settings.

\noindent
\begin{minipage}{\columnwidth}
\centering

\renewcommand{\arraystretch}{1.17}
\small
\setlength{\tabcolsep}{3pt}
\resizebox{\columnwidth}{!}{%
\begin{tabular}{@{}cccccc@{}}
\hline
\multirow{2}{*}{\textbf{Dataset}} & \multirow{2}{*}{\textbf{Pred. Len.}}
& \textbf{MOIRAI-small} & \textbf{+FSE}
& \textbf{TimesFM 2.5} & \textbf{+FSE} \\
\cmidrule(lr){3-3}\cmidrule(lr){4-4}\cmidrule(lr){5-5}\cmidrule(lr){6-6}
& & MSE\hspace{0.65em} MAE & MSE\hspace{0.65em} MAE & MSE\hspace{0.65em} MAE & MSE\hspace{0.65em} MAE \\
\hline
\multirow{5}{*}{ETTh1}
& 96  & 0.401\hspace{0.65em} 0.402 & 0.371\hspace{0.65em} 0.408
      & 0.414\hspace{0.65em} 0.404 & 0.386\hspace{0.65em} 0.426 \\
& 192 & 0.435\hspace{0.65em} 0.421 & 0.398\hspace{0.65em} 0.409
      & 0.465\hspace{0.65em} 0.434 & 0.413\hspace{0.65em} 0.442 \\
& 336 & 0.438\hspace{0.65em} 0.434 & 0.401\hspace{0.65em} 0.409
      & 0.503\hspace{0.65em} 0.456 & 0.440\hspace{0.65em} 0.461 \\
& 720 & 0.439\hspace{0.65em} 0.454 & 0.458\hspace{0.65em} 0.459
      & 0.511\hspace{0.65em} 0.481 & 0.519\hspace{0.65em} 0.524 \\

& \textbf{Avg.}
      & 0.428\hspace{0.65em} 0.427 & \textbf{0.407\hspace{0.65em} 0.421}
      & 0.473\hspace{0.65em} 0.443 & \textbf{0.440}\hspace{0.65em} 0.463 \\
\hline
\multirow{5}{*}{ETTh2}
& 96  & 0.297\hspace{0.65em} 0.336 & 0.263\hspace{0.65em} 0.329
      & 0.315\hspace{0.65em} 0.349 & 0.292\hspace{0.65em} 0.345 \\
& 192 & 0.368\hspace{0.65em} 0.381 & 0.319\hspace{0.65em} 0.372
      & 0.388\hspace{0.65em} 0.395 & 0.339\hspace{0.65em} 0.387 \\
& 336 & 0.370\hspace{0.65em} 0.393 & 0.311\hspace{0.65em} 0.373
      & 0.422\hspace{0.65em} 0.427 & 0.358\hspace{0.65em} 0.410 \\
& 720 & 0.411\hspace{0.65em} 0.426 & 0.401\hspace{0.65em} 0.434
      & 0.443\hspace{0.65em} 0.454 & 0.400\hspace{0.65em} 0.448 \\

& \textbf{Avg.}
      & 0.361\hspace{0.65em} 0.384 & \textbf{0.324\hspace{0.65em} 0.377}
      & 0.392\hspace{0.65em} 0.406 & \textbf{0.345\hspace{0.65em} 0.397} \\
\hline
\end{tabular}%
}
\captionof{table}{Per-horizon and average MSE and MAE for frozen time-series foundation models with and without FSE ($L=96$). Bold marks lower average errors within each model.}
\label{tab:foundation_fse}
\end{minipage}\par

\subsection{Extension to Time-Series Foundation Models}

Table~\ref{tab:foundation_fse} evaluates FSE trained on top of frozen MOIRAI-small~\cite{woo2024moirai} and TimesFM 2.5~\cite{das2024decoderonlyfoundationmodeltimeseries} at \(L=96\). On ETTh1/ETTh2, average MSE drops by 4.9\%/10.2\% for MOIRAI-small and 7.0\%/12.0\% for TimesFM 2.5. Average MAE also drops, except for TimesFM 2.5 on ETTh1 (0.443 to 0.463).

\begin{table*}[!t]
\begingroup
    \centering
    \begin{tabular}{l cc cc cc}
        \toprule
        \multirow{2}{*}{\textbf{Dataset}} & 
        \multicolumn{2}{c}{\textbf{TimesNet+FSE}} & 
        \multicolumn{2}{c}{\textbf{TimeMixer+FSE}} & 
        \multicolumn{2}{c}{\textbf{Dualformer+FSE}}\\
        
        \cmidrule(lr){2-3} \cmidrule(lr){4-5}\cmidrule(lr){6-7}
        
        & \textbf{MSE} & \textbf{MAE} & \textbf{MSE} & \textbf{MAE} & \textbf{MSE} & \textbf{MAE}\\
        \midrule

        ETTh1      & $0.408 \pm 0.010$ & $0.427 \pm 0.009$ & $0.401\pm 0.003$ & $0.418\pm 0.002$ & $0.396 \pm 0.014$ & $0.420\pm 0.003$\\
        ETTh2       & $0.336 \pm 0.011$ & $0.387 \pm 0.015$ & $0.333\pm 0.005$ & $0.381\pm 0.008$ & $0.324\pm 0.015$ & $0.373\pm 0.010$ \\
        ETTm1       & $0.350 \pm 0.009$ & $0.382 \pm 0.007$ & $0.342\pm 0.011$ & $0.367\pm 0.013$ & $0.339\pm 0.011$ & $0.361\pm 0.007$ \\
        ETTm2       & $0.262 \pm 0.004$ & $0.321 \pm 0.005$ & $0.243\pm 0.003$ & $0.302\pm 0.003$ &$0.238\pm 0.016$ & $0.292\pm 0.010$\\
        Electricity & $0.159 \pm 0.008$ & $0.256 \pm 0.006$ & $0.154\pm 0.005$ & $0.244\pm 0.004$ & $0.156\pm 0.007$ & $0.243\pm 0.005$\\
        Traffic     & $0.404 \pm 0.010$ & $0.281 \pm 0.013$ & $0.387\pm 0.012$ & $0.255\pm 0.008$ & $0.402\pm 0.009$ & $0.258\pm 0.005$\\
        Solar  & $0.210 \pm 0.011$ & $0.248 \pm 0.009$ & $0.195\pm 0.006$ & $0.230\pm 0.007$ & $0.198\pm 0.010$ & $0.238\pm 0.006$ \\
        
        \bottomrule
    \end{tabular}
    
    \begin{tabular}{l cc cc cc}
        \toprule
        \multirow{2}{*}{\textbf{Dataset}}& 
        \multicolumn{2}{c}{\textbf{DLinear+FSE}}& 
        \multicolumn{2}{c}{\textbf{PatchTST+FSE}} & 
        \multicolumn{2}{c}{\textbf{iTransformer+FSE}} \\
        
        \cmidrule(lr){2-3} \cmidrule(lr){4-5} \cmidrule(lr){6-7}
        
        & \textbf{MSE} & \textbf{MAE} & \textbf{MSE} & \textbf{MAE} & \textbf{MSE} & \textbf{MAE} \\
        \midrule

        ETTh1     &$0.430\pm 0.002$ & $0.445\pm 0.004$ & $0.409\pm 0.009$ & $0.418\pm 0.014$ & $0.421 \pm 0.009$ & $0.432 \pm 0.007$ \\
        ETTh2      & $0.445\pm 0.004$ &$0.443\pm 0.006$ & $0.340\pm 0.006$ & $0.388\pm 0.010$ & $0.345 \pm 0.014$ & $0.388 \pm 0.012$ \\
        ETTm1     & $0.355\pm 0.013$ & $0.375\pm 0.006$ & $0.351\pm 0.018$ & $0.383\pm 0.002$  & $0.369 \pm 0.011$ & $0.397 \pm 0.008$ \\
        ETTm2     & $0.275\pm 0.009$ & $0.330\pm 0.007$ & $0.242\pm 0.004$ & $0.310\pm 0.003$  & $0.265 \pm 0.006$ & $0.310 \pm 0.006$ \\
        Electricity  & $0.164\pm 0.012$ & $0.260\pm 0.008$ & $0.157\pm 0.011$ & $0.253\pm 0.008$ & $0.152 \pm 0.013$ & $0.250 \pm 0.008$ \\
        Traffic   & $0.435\pm 0.010$ & $0.296\pm 0.003$ & $0.400\pm 0.017$ & $0.278\pm 0.019$  & $0.413 \pm 0.010$ & $0.270 \pm 0.011$ \\
        Solar  & $0.320\pm 0.009$ & $0.383\pm 0.007$ & $0.223\pm 0.009$ & $0.269\pm 0.011$ & $0.230 \pm 0.012$ & $0.257 \pm 0.013$ \\
        
        \bottomrule
    \end{tabular}
    \caption{Multivariate long-term forecasting results with error bars (Mean ± STD) across three random seeds.}
    \label{tab:confidence_interval}
\par
\endgroup
\medskip
\begingroup
\centering
\setlength{\tabcolsep}{4pt}
\renewcommand{\arraystretch}{1.04}
\resizebox{\textwidth}{!}{%
\begin{tabular}{cccccccccc}
\toprule
\multirow{2}{*}{\textbf{Dataset}} & \multirow{2}{*}{\textbf{Pred. Len.}}
& \textbf{DLinear} & \textbf{+FSE} & \textbf{w/o cosine} & \textbf{w/o EMA}
& \textbf{correction-only} & \textbf{w/o reconstruction} & \textbf{w/o bound} & \textbf{Time domain} \\
&& MSE\enspace MAE & MSE\enspace MAE & MSE\enspace MAE & MSE\enspace MAE
& MSE\enspace MAE & MSE\enspace MAE & MSE\enspace MAE & MSE\enspace MAE \\
\midrule
\multirow{5}{*}{ETTh1}
& 96
& 0.375\enspace 0.396 & 0.367\enspace 0.394
& 0.370\enspace 0.396 & 0.372\enspace 0.399
& 0.372\enspace 0.396 & 0.374\enspace 0.395
& 0.384\enspace 0.404 & 0.410\enspace 0.430 \\
& 192
& 0.428\enspace 0.437 & 0.425\enspace 0.434
& 0.428\enspace 0.436 & 0.427\enspace 0.437
& 0.433\enspace 0.440 & 0.433\enspace 0.440
& 0.441\enspace 0.447 & 0.481\enspace 0.481 \\
& 336
& 0.448\enspace 0.449 & 0.440\enspace 0.449
& 0.443\enspace 0.452 & 0.446\enspace 0.454
& 0.467\enspace 0.463 & 0.467\enspace 0.463
& 0.482\enspace 0.473 & 0.482\enspace 0.471 \\
& 720
& 0.505\enspace 0.514 & 0.486\enspace 0.501
& 0.488\enspace 0.505 & 0.493\enspace 0.510
& 0.512\enspace 0.515 & 0.512\enspace 0.515
& 0.527\enspace 0.527 & 0.533\enspace 0.527 \\
& Avg
& 0.439\enspace 0.449 & 0.430\enspace 0.445
& 0.432\enspace 0.447 & 0.435\enspace 0.450
& 0.446\enspace 0.454 & 0.447\enspace 0.453
& 0.459\enspace 0.463 & 0.477\enspace 0.477 \\
\midrule
\multirow{5}{*}{ETTm1}
& 96
& 0.301\enspace 0.344 & 0.301\enspace 0.341
& 0.303\enspace 0.342 & 0.304\enspace 0.343
& 0.301\enspace 0.345 & 0.301\enspace 0.344
& 0.312\enspace 0.355 & 0.326\enspace 0.368 \\
& 192
& 0.336\enspace 0.366 & 0.332\enspace 0.361
& 0.333\enspace 0.362 & 0.335\enspace 0.363
& 0.333\enspace 0.367 & 0.333\enspace 0.367
& 0.342\enspace 0.375 & 0.364\enspace 0.398 \\
& 336
& 0.371\enspace 0.387 & 0.366\enspace 0.381
& 0.368\enspace 0.384 & 0.368\enspace 0.384
& 0.386\enspace 0.407 & 0.394\enspace 0.418
& 0.391\enspace 0.408 & 0.410\enspace 0.422 \\
& 720
& 0.426\enspace 0.422 & 0.420\enspace 0.415
& 0.423\enspace 0.418 & 0.422\enspace 0.418
& 0.447\enspace 0.445 & 0.454\enspace 0.453
& 0.440\enspace 0.433 & 0.487\enspace 0.473 \\
& Avg
& 0.359\enspace 0.380 & 0.355\enspace 0.375
& 0.357\enspace 0.377 & 0.357\enspace 0.377
& 0.367\enspace 0.391 & 0.371\enspace 0.396
& 0.371\enspace 0.393 & 0.397\enspace 0.415 \\
\midrule
\multirow{5}{*}{Electricity}
& 96
& 0.140\enspace 0.237 & 0.137\enspace 0.234
& 0.139\enspace 0.236 & 0.140\enspace 0.237
& 0.142\enspace 0.238 & 0.144\enspace 0.242
& 0.139\enspace 0.238 & 0.139\enspace 0.237 \\
& 192
& 0.154\enspace 0.250 & 0.153\enspace 0.248
& 0.156\enspace 0.249 & 0.156\enspace 0.250
& 0.155\enspace 0.254 & 0.163\enspace 0.258
& 0.155\enspace 0.261 & 0.161\enspace 0.253 \\
& 336
& 0.169\enspace 0.268 & 0.167\enspace 0.262
& 0.170\enspace 0.264 & 0.171\enspace 0.265
& 0.171\enspace 0.270 & 0.169\enspace 0.270
& 0.170\enspace 0.274 & 0.174\enspace 0.272 \\
& 720
& 0.204\enspace 0.300 & 0.200\enspace 0.294
& 0.204\enspace 0.296 & 0.207\enspace 0.298
& 0.209\enspace 0.302 & 0.206\enspace 0.303
& 0.207\enspace 0.307 & 0.206\enspace 0.302 \\
& Avg
& 0.167\enspace 0.264 & 0.164\enspace 0.260
& 0.167\enspace 0.261 & 0.169\enspace 0.263
& 0.169\enspace 0.266 & 0.171\enspace 0.268
& 0.168\enspace 0.270 & 0.170\enspace 0.266 \\
\midrule
\multirow{5}{*}{Traffic}
&96
& 0.412\enspace 0.286 & 0.414\enspace 0.286 & 0.419\enspace 0.293 & 0.417\enspace 0.291
& 0.463\enspace 0.317 & 0.412\enspace 0.286 & 0.512\enspace 0.330 & 0.479\enspace 0.319 \\
&192
& 0.424\enspace 0.291 & 0.423\enspace 0.289 & 0.427\enspace 0.292 & 0.430\enspace 0.295
& 0.429\enspace 0.297 & 0.462\enspace 0.308 & 0.445\enspace 0.296 & 0.429\enspace 0.298 \\
&336
& 0.438\enspace 0.299 & 0.437\enspace 0.296 & 0.443\enspace 0.301 & 0.445\enspace 0.300
& 0.444\enspace 0.304 & 0.444\enspace 0.301 & 0.457\enspace 0.305 & 0.449\enspace 0.304 \\
&720
& 0.467\enspace 0.317 & 0.465\enspace 0.311 & 0.467\enspace 0.313 & 0.468\enspace 0.313
& 0.472\enspace 0.327 & 0.495\enspace 0.326 & 0.470\enspace 0.319 & 0.492\enspace 0.319 \\
&Avg
& 0.435\enspace 0.298 & 0.435\enspace 0.296 & 0.439\enspace 0.300 & 0.440\enspace 0.300
& 0.452\enspace 0.311 & 0.453\enspace 0.305 & 0.471\enspace 0.313 & 0.462\enspace 0.310 \\
\bottomrule
\end{tabular}%
}
\caption{Per-horizon and average MSE and MAE for FSE ablations with DLinear as the backbone. Correction-only removes future guidance, and Time domain uses a matched-capacity time-domain corrector.}
\label{tab:full_ablation}
\par
\endgroup
\end{table*}

\subsection{Post-hoc Method Comparison}

Table~\ref{tab:full_posthoc} expands the comparison in Table~\ref{tab:average_performance_single} to individual prediction lengths. 

\begin{table*}[!t]
\centering
\small
\setlength{\tabcolsep}{2.25pt}
\renewcommand{\arraystretch}{0.96}
\begin{tabular}{@{}cc*{16}{c}@{}}
\toprule
 & & \multicolumn{8}{c}{\textbf{iTransformer}} & \multicolumn{8}{c}{\textbf{TimesNet}} \\
\cmidrule(lr){3-10}\cmidrule(lr){11-18}
\multirow{2}{*}{\textbf{Dataset}} & \multirow{2}{*}{\textbf{Pred. Len.}} & \multicolumn{2}{c}{\textbf{Base}} & \multicolumn{2}{c}{\textbf{+RevIN}} & \multicolumn{2}{c}{\textbf{+FAN}} & \multicolumn{2}{c}{\textbf{+FSE}} & \multicolumn{2}{c}{\textbf{Base}} & \multicolumn{2}{c}{\textbf{+RevIN}} & \multicolumn{2}{c}{\textbf{+FAN}} & \multicolumn{2}{c}{\textbf{+FSE}} \\
& & MSE & MAE & MSE & MAE & MSE & MAE & MSE & MAE & MSE & MAE & MSE & MAE & MSE & MAE & MSE & MAE \\
\midrule
\multirow{5}{*}{ETTh1} & 96 & 0.381 & 0.395 & 0.373 & 0.396 & 0.369 & 0.394 & \textbf{0.368} & \textbf{0.391} & 0.384 & 0.402 & 0.373 & 0.399 & 0.373 & 0.400 & \textbf{0.366} & \textbf{0.396} \\
 & 192 & 0.437 & 0.424 & 0.428 & 0.438 & 0.426 & 0.434 & \textbf{0.406} & \textbf{0.421} & 0.436 & 0.429 & 0.413 & 0.425 & 0.409 & 0.424 & \textbf{0.401} & \textbf{0.415} \\
 & 336 & 0.479 & 0.446 & \textbf{0.444} & \textbf{0.444} & 0.456 & 0.449 & 0.459 & \textbf{0.444} & 0.491 & 0.469 & 0.455 & 0.452 & 0.430 & 0.450 & \textbf{0.420} & \textbf{0.429} \\
 & 720 & 0.481 & \textbf{0.470} & 0.496 & 0.501 & 0.517 & 0.508 & \textbf{0.452} & \textbf{0.470} & 0.521 & 0.500 & 0.547 & 0.507 & 0.457 & \textbf{0.465} & \textbf{0.446} & 0.468 \\
 & \textbf{Avg.} & 0.445 & 0.434 & 0.435 & 0.445 & 0.442 & 0.446 & \textbf{0.421} & \textbf{0.432} & 0.458 & 0.450 & 0.447 & 0.446 & 0.417 & 0.435 & \textbf{0.408} & \textbf{0.427} \\
\midrule
\multirow{5}{*}{ETTh2} & 96 & 0.288 & 0.338 & 0.288 & 0.350 & 0.277 & 0.341 & \textbf{0.269} & \textbf{0.334} & 0.340 & 0.374 & 0.281 & 0.341 & 0.271 & \textbf{0.335} & \textbf{0.269} & 0.336 \\
 & 192 & 0.374 & 0.390 & \textbf{0.352} & 0.392 & 0.379 & 0.408 & 0.374 & \textbf{0.376} & 0.402 & 0.414 & 0.395 & 0.407 & 0.336 & \textbf{0.376} & \textbf{0.333} & 0.378 \\
 & 336 & 0.415 & 0.426 & 0.379 & \textbf{0.413} & 0.365 & 0.417 & \textbf{0.358} & \textbf{0.413} & 0.452 & 0.452 & 0.434 & 0.456 & 0.374 & 0.414 & \textbf{0.359} & \textbf{0.405} \\
 & 720 & 0.420 & 0.440 & 0.412 & 0.450 & 0.409 & 0.448 & \textbf{0.377} & \textbf{0.429} & 0.462 & 0.468 & 0.440 & 0.461 & 0.394 & 0.440 & \textbf{0.384} & \textbf{0.428} \\
 & \textbf{Avg.} & 0.374 & 0.399 & 0.358 & 0.401 & 0.358 & 0.404 & \textbf{0.345} & \textbf{0.388} & 0.414 & 0.427 & 0.388 & 0.416 & 0.344 & 0.391 & \textbf{0.336} & \textbf{0.387} \\
\midrule
\multirow{5}{*}{ETTm1} & 96 & 0.334 & 0.368 & 0.315 & 0.368 & \textbf{0.308} & 0.352 & 0.310 & \textbf{0.350} & 0.338 & 0.375 & 0.295 & \textbf{0.346} & 0.298 & 0.349 & \textbf{0.293} & 0.349 \\
 & 192 & 0.377 & 0.391 & 0.360 & 0.395 & 0.346 & 0.388 & \textbf{0.345} & \textbf{0.386} & \textbf{0.327} & 0.387 & 0.330 & \textbf{0.369} & 0.340 & 0.371 & 0.332 & 0.373 \\
 & 336 & 0.426 & 0.420 & 0.400 & 0.421 & 0.380 & \textbf{0.405} & \textbf{0.379} & 0.407 & 0.410 & 0.411 & 0.373 & 0.398 & 0.370 & 0.403 & \textbf{0.360} & \textbf{0.394} \\
 & 720 & 0.491 & 0.459 & 0.461 & \textbf{0.443} & 0.499 & 0.489 & \textbf{0.443} & 0.444 & 0.478 & 0.450 & 0.477 & 0.464 & 0.437 & 0.439 & \textbf{0.413} & \textbf{0.411} \\
 & \textbf{Avg.} & 0.407 & 0.410 & 0.384 & 0.407 & 0.383 & 0.409 & \textbf{0.369} & \textbf{0.397} & 0.388 & 0.406 & 0.369 & 0.394 & 0.361 & 0.391 & \textbf{0.350} & \textbf{0.382} \\
\midrule
\multirow{5}{*}{ETTm2} & 96 & 0.180 & 0.264 & 0.176 & 0.268 & 0.163 & 0.254 & \textbf{0.160} & \textbf{0.241} & 0.187 & 0.267 & 0.174 & 0.258 & 0.172 & 0.257 & \textbf{0.162} & \textbf{0.254} \\
 & 192 & 0.250 & 0.309 & 0.234 & 0.307 & \textbf{0.217} & 0.291 & 0.223 & \textbf{0.284} & 0.249 & 0.309 & 0.248 & 0.317 & 0.262 & 0.317 & \textbf{0.216} & \textbf{0.290} \\
 & 336 & 0.311 & 0.348 & 0.295 & 0.347 & \textbf{0.273} & 0.331 & 0.286 & \textbf{0.328} & 0.321 & 0.351 & 0.317 & 0.361 & 0.282 & \textbf{0.334} & \textbf{0.278} & 0.336 \\
 & 720 & 0.412 & 0.407 & 0.418 & 0.426 & 0.485 & 0.453 & \textbf{0.389} & \textbf{0.388} & 0.408 & \textbf{0.403} & 0.409 & 0.431 & \textbf{0.389} & 0.411 & 0.390 & 0.405 \\
 & \textbf{Avg.} & 0.288 & 0.332 & 0.281 & 0.337 & 0.285 & 0.332 & \textbf{0.265} & \textbf{0.310} & 0.291 & 0.333 & 0.287 & 0.342 & 0.276 & 0.330 & \textbf{0.262} & \textbf{0.321} \\
\midrule
 & & \multicolumn{8}{c}{\textbf{TimeMixer}} & \multicolumn{8}{c}{\textbf{PatchTST}} \\
\cmidrule(lr){3-10}\cmidrule(lr){11-18}
\multirow{2}{*}{\textbf{Dataset}} & \multirow{2}{*}{\textbf{Pred. Len.}} & \multicolumn{2}{c}{\textbf{Base}} & \multicolumn{2}{c}{\textbf{+RevIN}} & \multicolumn{2}{c}{\textbf{+FAN}} & \multicolumn{2}{c}{\textbf{+FSE}} & \multicolumn{2}{c}{\textbf{Base}} & \multicolumn{2}{c}{\textbf{+RevIN}} & \multicolumn{2}{c}{\textbf{+FAN}} & \multicolumn{2}{c}{\textbf{+FSE}} \\
& & MSE & MAE & MSE & MAE & MSE & MAE & MSE & MAE & MSE & MAE & MSE & MAE & MSE & MAE & MSE & MAE \\
\midrule
\multirow{5}{*}{ETTh1} & 96 & 0.361 & 0.390 & \textbf{0.358} & 0.398 & 0.367 & 0.387 & 0.361 & \textbf{0.383} & 0.373 & 0.402 & \textbf{0.368} & \textbf{0.386} & 0.373 & 0.397 & 0.369 & 0.398 \\
 & 192 & 0.409 & 0.414 & \textbf{0.382} & 0.412 & 0.410 & 0.415 & 0.387 & \textbf{0.409} & 0.411 & 0.428 & 0.412 & \textbf{0.413} & 0.417 & 0.422 & \textbf{0.407} & 0.422 \\
 & 336 & 0.430 & \textbf{0.429} & \textbf{0.421} & 0.432 & 0.434 & \textbf{0.429} & 0.424 & 0.430 & \textbf{0.432} & 0.445 & 0.433 & \textbf{0.426} & 0.448 & 0.443 & 0.443 & 0.449 \\
 & 720 & 0.445 & 0.460 & 0.444 & 0.460 & 0.434 & 0.452 & \textbf{0.433} & \textbf{0.451} & 0.455 & 0.473 & 0.425 & 0.443 & 0.484 & 0.500 & \textbf{0.415} & \textbf{0.404} \\
 & \textbf{Avg.} & 0.411 & 0.423 & \textbf{0.401} & 0.426 & 0.411 & 0.421 & \textbf{0.401} & \textbf{0.418} & 0.418 & 0.437 & 0.410 & \textbf{0.417} & 0.431 & 0.441 & \textbf{0.409} & 0.418 \\
\midrule
\multirow{5}{*}{ETTh2} & 96 & 0.271 & 0.330 & \textbf{0.265} & 0.334 & 0.275 & 0.337 & 0.266 & \textbf{0.328} & 0.275 & 0.338 & 0.296 & 0.353 & 0.284 & \textbf{0.322} & \textbf{0.271} & 0.335 \\
 & 192 & \textbf{0.317} & 0.402 & 0.323 & \textbf{0.363} & 0.339 & 0.379 & 0.330 & 0.372 & 0.338 & 0.380 & 0.369 & 0.400 & 0.372 & 0.382 & \textbf{0.336} & \textbf{0.376} \\
 & 336 & 0.332 & 0.396 & 0.368 & 0.401 & \textbf{0.331} & \textbf{0.380} & 0.360 & 0.399 & 0.366 & 0.403 & 0.381 & 0.410 & 0.408 & 0.419 & \textbf{0.357} & \textbf{0.401} \\
 & 720 & \textbf{0.342} & \textbf{0.408} & 0.416 & 0.440 & 0.421 & 0.494 & 0.374 & 0.423 & \textbf{0.391} & 0.431 & 0.401 & 0.428 & 0.409 & \textbf{0.421} & 0.394 & 0.440 \\
 & \textbf{Avg.} & \textbf{0.316} & 0.384 & 0.343 & 0.385 & 0.342 & 0.398 & 0.333 & \textbf{0.381} & 0.343 & 0.388 & 0.362 & 0.398 & 0.368 & \textbf{0.386} & \textbf{0.340} & 0.388 \\
\midrule
\multirow{5}{*}{ETTm1} & 96 & 0.291 & 0.340 & 0.289 & 0.343 & \textbf{0.288} & \textbf{0.329} & \textbf{0.288} & 0.330 & \textbf{0.288} & \textbf{0.341} & 0.294 & 0.343 & 0.301 & 0.345 & 0.293 & 0.348 \\
 & 192 & 0.327 & 0.365 & 0.336 & 0.371 & \textbf{0.318} & \textbf{0.342} & 0.322 & 0.356 & \textbf{0.332} & 0.370 & 0.339 & 0.373 & 0.336 & \textbf{0.366} & 0.335 & 0.376 \\
 & 336 & 0.360 & 0.381 & 0.367 & 0.392 & \textbf{0.349} & \textbf{0.351} & 0.350 & 0.375 & \textbf{0.362} & 0.392 & 0.372 & 0.393 & 0.372 & \textbf{0.389} & \textbf{0.362} & 0.397 \\
 & 720 & 0.415 & 0.417 & 0.419 & 0.425 & 0.452 & 0.432 & \textbf{0.407} & \textbf{0.406} & 0.416 & 0.419 & 0.424 & 0.428 & 0.427 & 0.423 & \textbf{0.413} & \textbf{0.411} \\
 & \textbf{Avg.} & 0.348 & 0.376 & 0.353 & 0.383 & 0.352 & \textbf{0.364} & \textbf{0.342} & 0.367 & \textbf{0.350} & \textbf{0.381} & 0.357 & 0.384 & 0.359 & \textbf{0.381} & 0.351 & 0.383 \\
\midrule
\multirow{5}{*}{ETTm2} & 96 & 0.164 & 0.254 & 0.164 & 0.254 & 0.169 & 0.264 & \textbf{0.155} & \textbf{0.241} & 0.162 & 0.254 & 0.161 & \textbf{0.242} & 0.164 & 0.251 & \textbf{0.152} & 0.257 \\
 & 192 & 0.223 & 0.295 & 0.224 & 0.294 & 0.279 & 0.395 & \textbf{0.210} & \textbf{0.280} & 0.217 & 0.293 & 0.218 & 0.293 & 0.218 & 0.289 & \textbf{0.211} & \textbf{0.287} \\
 & 336 & 0.279 & 0.330 & 0.278 & 0.330 & 0.306 & 0.350 & \textbf{0.263} & \textbf{0.315} & 0.267 & 0.326 & 0.276 & 0.330 & 0.266 & \textbf{0.322} & \textbf{0.262} & 0.324 \\
 & 720 & 0.359 & 0.383 & 0.367 & 0.385 & 0.431 & 0.441 & \textbf{0.343} & \textbf{0.372} & 0.353 & 0.382 & 0.345 & 0.382 & 0.361 & 0.384 & \textbf{0.342} & \textbf{0.372} \\
 & \textbf{Avg.} & 0.256 & 0.316 & 0.258 & 0.316 & 0.296 & 0.363 & \textbf{0.243} & \textbf{0.302} & 0.250 & 0.314 & 0.250 & 0.312 & 0.252 & 0.312 & \textbf{0.242} & \textbf{0.310} \\
\bottomrule
\end{tabular}
\caption{Per-horizon and average MSE and MAE for post-hoc methods on the four ETT datasets ($L=96$). Bold marks the minimum of each metric within the same backbone, dataset, and horizon or average, including ties.}
\label{tab:full_posthoc}
\end{table*}

\subsection{Forecasting Results Across Backbones}

Table~\ref{tab:full_main_results} gives the per-horizon results for the six backbones in Table~\ref{tab:fse_all_models}.

\begin{table*}[t]
\centering
\renewcommand{\arraystretch}{0.86}
\resizebox{.85\textwidth}{!}{
\begin{tabular}{c|c|cc|cc|cc|cc|cc|cc}
\toprule

\multicolumn{2}{c|}{\textbf{Models}} 
& \multicolumn{2}{c|}{\textbf{TimesNet}} 
& \multicolumn{2}{c|}{\textbf{+FSE}} 
& \multicolumn{2}{c|}{\textbf{TimeMixer}} 
& \multicolumn{2}{c|}{\textbf{+FSE}} 
& \multicolumn{2}{c|}{\textbf{Dualformer}} 
& \multicolumn{2}{c}{\textbf{+FSE}} \\

\multicolumn{2}{c|}{\textbf{Metric}} 
& MSE & MAE 
& MSE & MAE 
& MSE & MAE 
& MSE & MAE 
& MSE & MAE 
& MSE & MAE \\

\midrule

\multirow{4}{*}{\rotatebox{90}{\emph{ETTh1}}}
& 96  & 0.384 & 0.402 & \textbf{0.366} & \textbf{0.396} & 0.361 & 0.390 & 0.361 & \textbf{0.383} & 0.359 & 0.386 & \textbf{0.355} & 0.393 \\
& 192 & 0.436 & 0.429 & \textbf{0.401} & \textbf{0.415} & 0.409 & 0.414 & \textbf{0.387} & \textbf{0.409} & 0.398 & 0.412 & \textbf{0.394} & \textbf{0.411} \\
& 336 & 0.491 & 0.469 & \textbf{0.420} & \textbf{0.429} & 0.430 & 0.429 & \textbf{0.424} & 0.430 & 0.432 & 0.431 & \textbf{0.430} & \textbf{0.429} \\
& 720 & 0.521 & 0.500 & \textbf{0.446} & \textbf{0.468} & 0.445 & 0.460 & \textbf{0.433} & \textbf{0.451} & 0.438 & 0.453 & \textbf{0.405} & \textbf{0.448} \\
\midrule

\multirow{4}{*}{\rotatebox{90}{\emph{ETTh2}}}
& 96  & 0.340 & 0.374 & \textbf{0.269} & \textbf{0.336} & 0.271 & 0.330 & \textbf{0.266} & \textbf{0.328} & 0.268 & 0.329 & \textbf{0.259} & 0.330 \\
& 192 & 0.402 & 0.414 & \textbf{0.333} & \textbf{0.378} & 0.317 & 0.402 & 0.330 & \textbf{0.372} & 0.332 & 0.370 & \textbf{0.322} & \textbf{0.362} \\
& 336 & 0.452 & 0.452 & \textbf{0.359} & \textbf{0.405} & 0.332 & 0.396 & 0.360 & 0.399 & 0.359 & 0.393 & \textbf{0.349} & \textbf{0.386} \\
& 720 & 0.462 & 0.468 & \textbf{0.384} & \textbf{0.428} & 0.342 & 0.408 & 0.374 & 0.423 & 0.381 & 0.417 & \textbf{0.367} & \textbf{0.415} \\
\midrule

\multirow{4}{*}{\rotatebox{90}{\emph{ETTm1}}}
& 96  & 0.338 & 0.375 & \textbf{0.293} & \textbf{0.349} & 0.291 & 0.340 & \textbf{0.288} & \textbf{0.330} & 0.283 & 0.330 & 0.285 & \textbf{0.329} \\
& 192 & 0.327 & 0.387 & 0.332 & \textbf{0.373} & 0.327 & 0.365 & \textbf{0.322} & \textbf{0.356} & 0.326 & 0.355 & \textbf{0.318} & \textbf{0.347} \\
& 336 & 0.410 & 0.411 & \textbf{0.360} & \textbf{0.394} & 0.360 & 0.381 & \textbf{0.350} & \textbf{0.375} & 0.359 & 0.377 & \textbf{0.349} & \textbf{0.368} \\
& 720 & 0.478 & 0.450 & \textbf{0.413} & \textbf{0.411} & 0.415 & 0.417 & \textbf{0.407} & \textbf{0.406} & 0.412 & 0.408 & \textbf{0.402} & \textbf{0.399} \\
\midrule

\multirow{4}{*}{\rotatebox{90}{\emph{ETTm2}}}
& 96  & 0.187 & 0.267 & \textbf{0.162} & \textbf{0.254} & 0.164 & 0.254 & \textbf{0.155} & \textbf{0.241} & 0.157 & 0.244 & \textbf{0.152} & \textbf{0.235} \\
& 192 & 0.249 & 0.309 & \textbf{0.216} & \textbf{0.290} & 0.223 & 0.295 & \textbf{0.210} & \textbf{0.280} & 0.213 & 0.282 & \textbf{0.205} & \textbf{0.268} \\
& 336 & 0.321 & 0.351 & \textbf{0.278} & \textbf{0.336} & 0.279 & 0.330 & \textbf{0.263} & \textbf{0.315} & 0.264 & 0.318 & \textbf{0.260} & \textbf{0.310} \\
& 720 & 0.408 & 0.403 & \textbf{0.390} & 0.405 & 0.359 & 0.383 & \textbf{0.343} & \textbf{0.372} & 0.345 & 0.371 & \textbf{0.335} & \textbf{0.354} \\
\midrule

\multirow{4}{*}{\rotatebox{90}{\emph{Electricity}}}
& 96  & 0.168 & 0.272 & \textbf{0.128} & \textbf{0.223} & 0.135 & 0.222 & \textbf{0.126} & \textbf{0.216} & 0.128 & 0.221 & \textbf{0.122} & \textbf{0.209} \\
& 192 & 0.184 & 0.289 & \textbf{0.148} & \textbf{0.244} & 0.147 & 0.235 & \textbf{0.143} & \textbf{0.233} & 0.145 & 0.237 & \textbf{0.142} & \textbf{0.228} \\
& 336 & 0.198 & 0.300 & \textbf{0.164} & \textbf{0.262} & 0.164 & 0.245 & \textbf{0.158} & 0.249 & 0.160 & 0.253 & 0.162 & \textbf{0.251} \\
& 720 & 0.220 & 0.320 & \textbf{0.196} & \textbf{0.294} & 0.212 & 0.310 & \textbf{0.189} & \textbf{0.278} & 0.197 & 0.285 & 0.199 & \textbf{0.283} \\
\midrule

\multirow{4}{*}{\rotatebox{90}{\emph{Traffic}}}
& 96  & 0.593 & 0.321 & \textbf{0.363} & \textbf{0.265} & 0.366 & 0.259 & \textbf{0.349} & \textbf{0.225} & 0.377 & 0.256 & \textbf{0.368} & \textbf{0.233} \\
& 192 & 0.617 & 0.336 & \textbf{0.385} & \textbf{0.273} & 0.381 & 0.265 & 0.381 & \textbf{0.261} & 0.392 & 0.263 & \textbf{0.384} & \textbf{0.250} \\
& 336 & 0.629 & 0.336 & \textbf{0.419} & \textbf{0.277} & 0.397 & 0.269 & 0.397 & 0.270 & 0.406 & 0.266 & 0.408 & \textbf{0.264} \\
& 720 & 0.640 & 0.350 & \textbf{0.449} & \textbf{0.308} & 0.429 & 0.292 & \textbf{0.422} & \textbf{0.263} & 0.447 & 0.287 & 0.449 & \textbf{0.285} \\
\midrule

\multirow{4}{*}{\rotatebox{90}{\emph{Solar}}}
& 96  & 0.219 & 0.314 & \textbf{0.188} & \textbf{0.232} & 0.167 & 0.220 & 0.172 & \textbf{0.200} & 0.175 & 0.228 & \textbf{0.168} & \textbf{0.226} \\
& 192 & 0.231 & 0.322 & \textbf{0.212} & \textbf{0.248} & 0.187 & 0.249 & 0.189 & \textbf{0.213} & 0.186 & 0.235 & 0.198 & \textbf{0.233} \\
& 336 & 0.246 & 0.337 & \textbf{0.218} & \textbf{0.250} & 0.200 & 0.258 & 0.204 & \textbf{0.249} & 0.200 & 0.246 & 0.215 & 0.247 \\
& 720 & 0.280 & 0.363 & \textbf{0.223} & \textbf{0.261} & 0.215 & 0.250 & 0.215 & 0.256 & 0.203 & 0.249 & 0.210 & \textbf{0.244} \\
\bottomrule

\end{tabular}
}
\resizebox{.85\textwidth}{!}{
\begin{tabular}{c|c|cc|cc|cc|cc|cc|cc}
\toprule
\multicolumn{2}{c|}{\textbf{Models}}
& \multicolumn{2}{c|}{\textbf{DLinear}}
& \multicolumn{2}{c|}{\textbf{+FSE}}
& \multicolumn{2}{c|}{\textbf{PatchTST}}
& \multicolumn{2}{c|}{\textbf{+FSE}}
& \multicolumn{2}{c|}{\textbf{iTransformer}}
& \multicolumn{2}{c}{\textbf{+FSE}} \\

\multicolumn{2}{c|}{\textbf{Metric}}
& MSE & MAE & MSE & MAE & MSE & MAE & MSE & MAE & MSE & MAE & MSE & MAE \\
\midrule

\multirow{4}{*}{\rotatebox{90}{\emph{ETTh1}}}
& 96  & 0.375 & 0.396 & \textbf{0.367} & \textbf{0.394} & 0.373 & 0.402 & \textbf{0.369} & \textbf{0.398} & 0.381 & 0.395 & \textbf{0.368} & \textbf{0.391} \\
& 192 & 0.428 & 0.437 & \textbf{0.425} & \textbf{0.434} & 0.411 & 0.428 & \textbf{0.407} & \textbf{0.422} & 0.437 & 0.424 & \textbf{0.406} & \textbf{0.421} \\
& 336 & 0.448 & 0.449 & \textbf{0.440} & 0.449 & 0.432 & 0.445 & 0.443 & 0.449 & 0.479 & 0.446 & \textbf{0.459} & \textbf{0.444} \\
& 720 & 0.505 & 0.514 & \textbf{0.486} & \textbf{0.501} & 0.455 & 0.473 & \textbf{0.415} & \textbf{0.404} & 0.481 & 0.470 & \textbf{0.452} & 0.470 \\
\midrule

\multirow{4}{*}{\rotatebox{90}{\emph{ETTh2}}}
& 96  & 0.296 & 0.360 & \textbf{0.290} & \textbf{0.351} & 0.275 & 0.338 & \textbf{0.271} & \textbf{0.335} & 0.288 & 0.338 & \textbf{0.269} & \textbf{0.334} \\
& 192 & 0.391 & 0.423 & \textbf{0.370} & \textbf{0.401} & 0.338 & 0.380 & \textbf{0.336} & \textbf{0.376} & 0.374 & 0.390 & 0.374 & \textbf{0.376} \\
& 336 & 0.445 & 0.460 & \textbf{0.417} & \textbf{0.434} & 0.366 & 0.403 & \textbf{0.357} & \textbf{0.401} & 0.415 & 0.426 & \textbf{0.358} & \textbf{0.413} \\
& 720 & 0.700 & 0.592 & 0.702 & \textbf{0.586} & 0.391 & 0.431 & 0.394 & 0.440 & 0.420 & 0.440 & \textbf{0.377} & \textbf{0.429} \\
\midrule

\multirow{4}{*}{\rotatebox{90}{\emph{ETTm1}}}
& 96  & 0.301 & 0.344 & 0.301 & \textbf{0.341} & 0.288 & 0.341 & 0.293 & 0.348 & 0.334 & 0.368 & \textbf{0.310} & \textbf{0.350} \\
& 192 & 0.336 & 0.366 & \textbf{0.332} & \textbf{0.361} & 0.332 & 0.370 & 0.335 & 0.376 & 0.377 & 0.391 & \textbf{0.345} & \textbf{0.386} \\
& 336 & 0.371 & 0.387 & \textbf{0.366} & \textbf{0.381} & 0.362 & 0.392 & 0.362 & 0.397 & 0.426 & 0.420 & \textbf{0.379} & \textbf{0.407} \\
& 720 & 0.426 & 0.422 & \textbf{0.420} & \textbf{0.415} & 0.416 & 0.419 & \textbf{0.413} & \textbf{0.411} & 0.491 & 0.459 & \textbf{0.443} & \textbf{0.444} \\
\midrule

\multirow{4}{*}{\rotatebox{90}{\emph{ETTm2}}}
& 96  & 0.170 & 0.264 & \textbf{0.168} & \textbf{0.257} & 0.162 & 0.254 & \textbf{0.152} & 0.257 & 0.180 & 0.264 & \textbf{0.160} & \textbf{0.241} \\
& 192 & 0.233 & 0.311 & \textbf{0.227} & \textbf{0.294} & 0.217 & 0.293 & \textbf{0.211} & \textbf{0.287} & 0.250 & 0.309 & \textbf{0.223} & \textbf{0.284} \\
& 336 & 0.300 & 0.358 & \textbf{0.293} & \textbf{0.346} & 0.267 & 0.326 & \textbf{0.262} & \textbf{0.324} & 0.311 & 0.348 & \textbf{0.286} & \textbf{0.328} \\
& 720 & 0.422 & 0.439 & \textbf{0.410} & \textbf{0.423} & 0.353 & 0.382 & \textbf{0.342} & \textbf{0.372} & 0.412 & 0.407 & \textbf{0.389} & \textbf{0.388} \\
\midrule

\multirow{4}{*}{\rotatebox{90}{\emph{Electricity}}}
& 96  & 0.140 & 0.237 & \textbf{0.137} & \textbf{0.234} & 0.129 & 0.223 & \textbf{0.128} & 0.224 & 0.131 & 0.227 & \textbf{0.124} & \textbf{0.220} \\
& 192 & 0.154 & 0.250 & \textbf{0.153} & \textbf{0.248} & 0.149 & 0.243 & \textbf{0.145} & \textbf{0.240} & 0.148 & 0.244 & \textbf{0.146} & \textbf{0.242} \\
& 336 & 0.169 & 0.268 & \textbf{0.167} & \textbf{0.262} & 0.165 & 0.260 & \textbf{0.160} & \textbf{0.258} & 0.164 & 0.262 & \textbf{0.156} & \textbf{0.256} \\
& 720 & 0.204 & 0.300 & \textbf{0.200} & \textbf{0.294} & 0.200 & 0.292 & \textbf{0.194} & \textbf{0.291} & 0.200 & 0.295 & \textbf{0.182} & \textbf{0.281} \\
\midrule

\multirow{4}{*}{\rotatebox{90}{\emph{Traffic}}}
& 96  & 0.412 & 0.286 & 0.414 & 0.286 & 0.383 & 0.272 & \textbf{0.378} & \textbf{0.266} & 0.395 & 0.268 & \textbf{0.365} & \textbf{0.241} \\
& 192 & 0.424 & 0.291 & \textbf{0.423} & \textbf{0.289} & 0.380 & 0.259 & \textbf{0.368} & \textbf{0.246} & 0.417 & 0.276 & \textbf{0.396} & \textbf{0.258} \\
& 336 & 0.438 & 0.299 & \textbf{0.437} & \textbf{0.296} & 0.410 & 0.285 & \textbf{0.409} & 0.289 & 0.433 & 0.283 & \textbf{0.432} & \textbf{0.272} \\
& 720 & 0.467 & 0.317 & \textbf{0.465} & \textbf{0.311} & 0.454 & 0.313 & \textbf{0.445} & \textbf{0.309} & 0.467 & 0.302 & \textbf{0.459} & 0.307 \\
\midrule

\multirow{4}{*}{\rotatebox{90}{\emph{Solar}}}
& 96  & 0.289 & 0.337 & \textbf{0.282} & \textbf{0.330} & 0.224 & 0.278 & \textbf{0.207} & \textbf{0.252} & 0.203 & 0.237 & 0.203 & \textbf{0.233} \\
& 192 & 0.319 & 0.397 & \textbf{0.313} & \textbf{0.392} & 0.253 & 0.298 & \textbf{0.229} & \textbf{0.274} & 0.233 & 0.261 & \textbf{0.226} & \textbf{0.260} \\
& 336 & 0.352 & 0.415 & \textbf{0.342} & \textbf{0.407} & 0.273 & 0.306 & \textbf{0.215} & \textbf{0.265} & 0.248 & 0.273 & \textbf{0.246} & \textbf{0.265} \\
& 720 & 0.356 & 0.412 & \textbf{0.342} & \textbf{0.404} & 0.272 & 0.308 & \textbf{0.240} & \textbf{0.285} & 0.249 & 0.275 & \textbf{0.244} & \textbf{0.270} \\

\bottomrule
\end{tabular}
}
\caption{Full results of performance comparison between original models and models enhanced with FSE.}
\label{tab:full_main_results}
\end{table*}

\section{Visualizations}

\begin{figure*}[h]
    \centering
    \includegraphics[width=\linewidth]{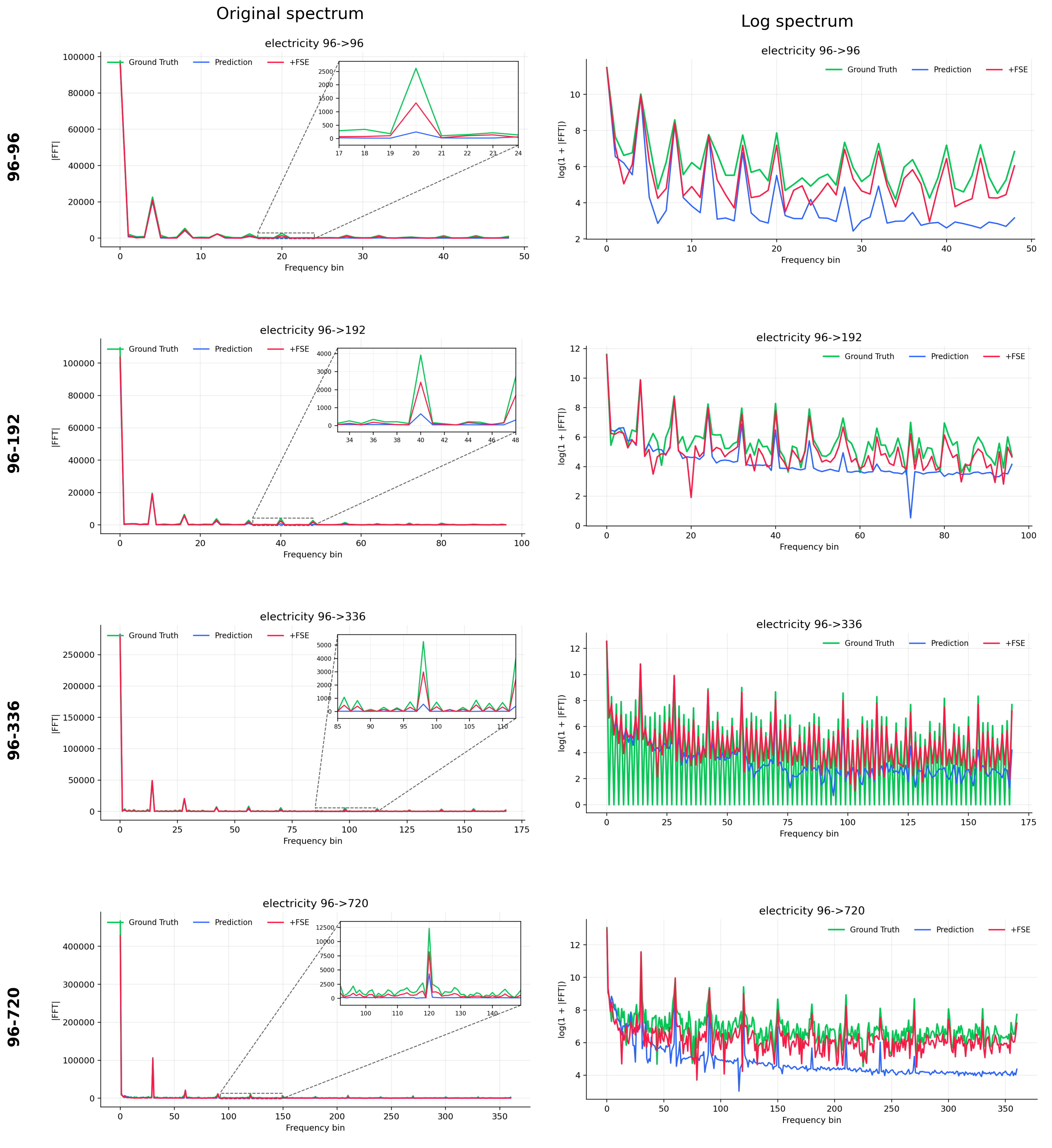}
    \caption{Spectral comparison on the Electricity dataset across different prediction lengths under the same lookback window size $L=96$. The left column shows the original FFT magnitude spectrum with zoomed mid-to-high frequency regions, and the right column shows the corresponding log-scaled spectrum. Compared with the baseline prediction, FSE better recovers attenuated mid- and high-frequency components and produces spectra closer to the ground truth with an adaptively adjusted correction bound.}
    \label{fig:ecl_spectrum}
\end{figure*}

\begin{figure*}[h]
    \centering
    \includegraphics[width=\linewidth]{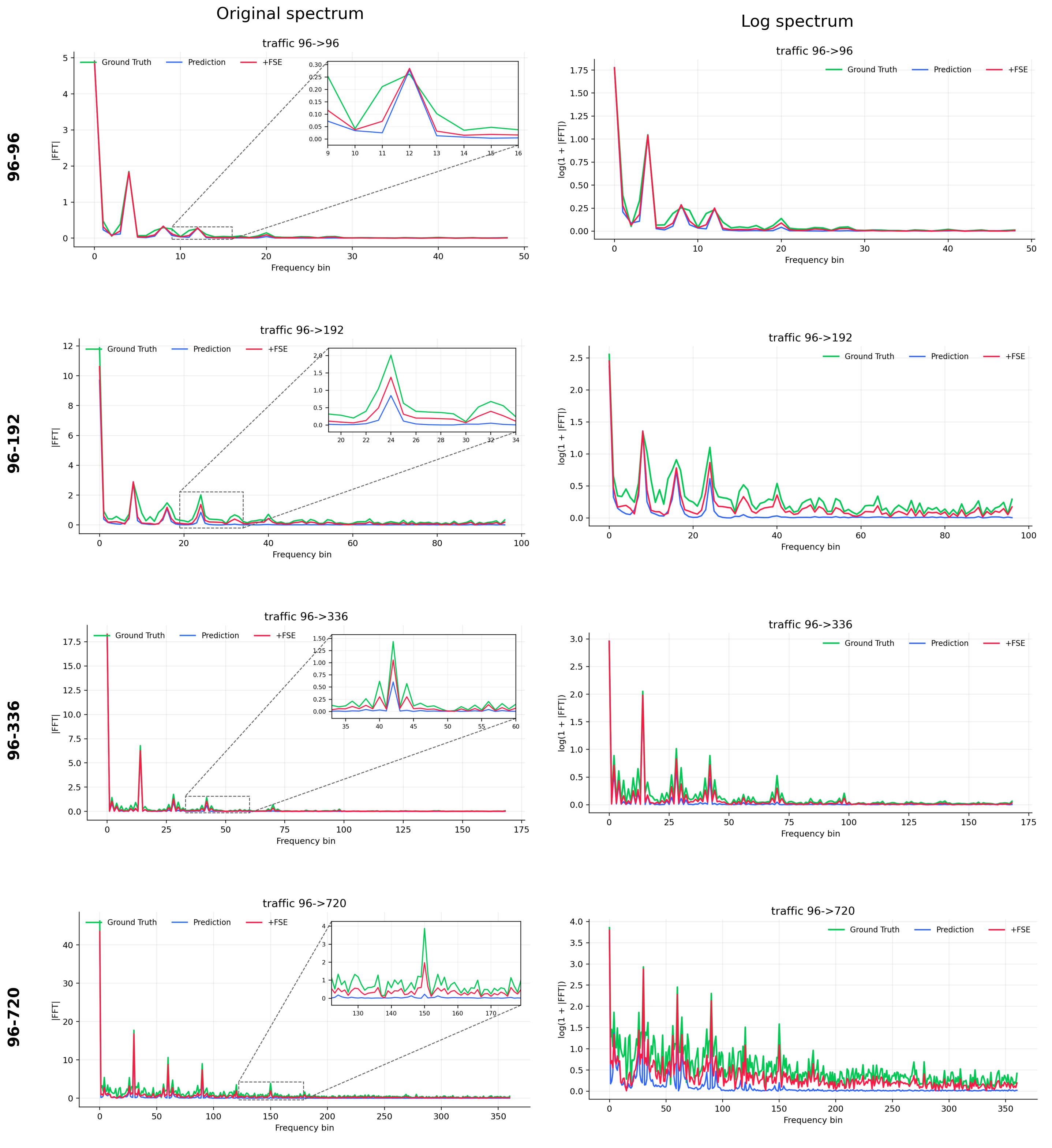}
    \caption{Spectral comparison on the Traffic dataset across different prediction lengths under the same lookback window size $L=96$. The left column shows the original FFT magnitude spectrum with zoomed mid-to-high frequency regions, and the right column shows the corresponding log-scaled spectrum. Compared with the baseline prediction, FSE better recovers attenuated mid- and high-frequency components and produces spectra closer to the ground truth with an adaptively adjusted correction bound.}
    \label{fig:tra_spectrum}
\end{figure*}

We provide additional qualitative results to better understand the effect of FSE. Figures~\ref{fig:ecl_spectrum} and~\ref{fig:tra_spectrum} compared the frequency spectra of the ground truth, baseline prediction, and FSE-enhanced prediction on Electricity and Traffic datasets. For each dataset, we show both the original spectrum and the log-scaled spectrum across different prediction lengths. The baseline prediction tends to underestimate several mid- and high-frequency components, which is consistent with the over-smoothing phenomenon observed in long-term forecasting. After applying FSE, the corrected spectrum becomes closer to the ground truth, especially around local spectral peaks and high-frequency regions. This supports the claim that FSE does not simply improve aggregate errors, but directly mitigates spectral attenuation.

\begin{figure*}[h]
    \centering
    \includegraphics[width=\linewidth]{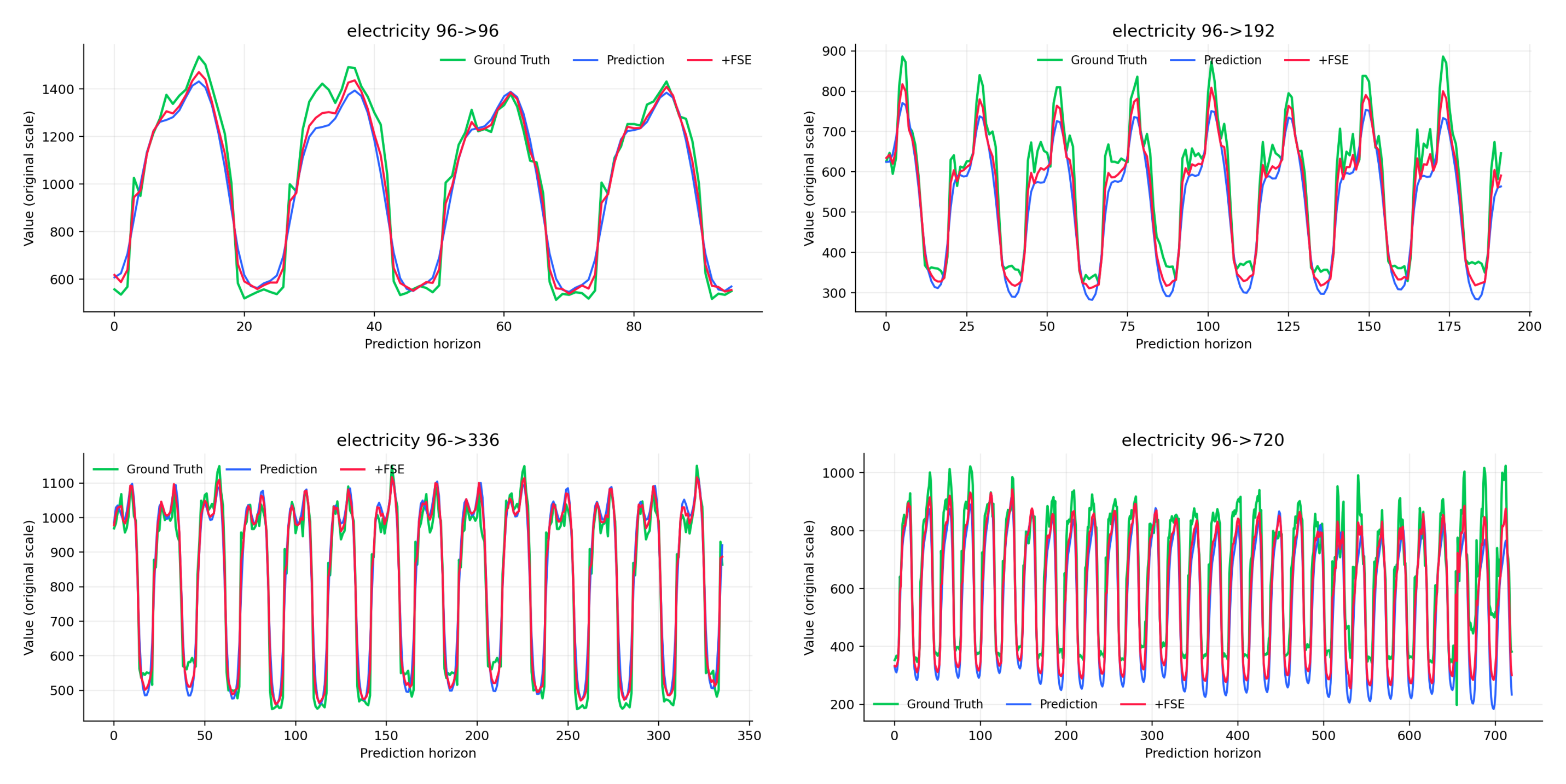}
    \caption{Time-domain forecasting visualization on the Electricity dataset across different prediction lengths.}
    \label{fig:ecl_time}
\end{figure*}

\begin{figure*}[h]
    \centering
    \includegraphics[width=\linewidth]{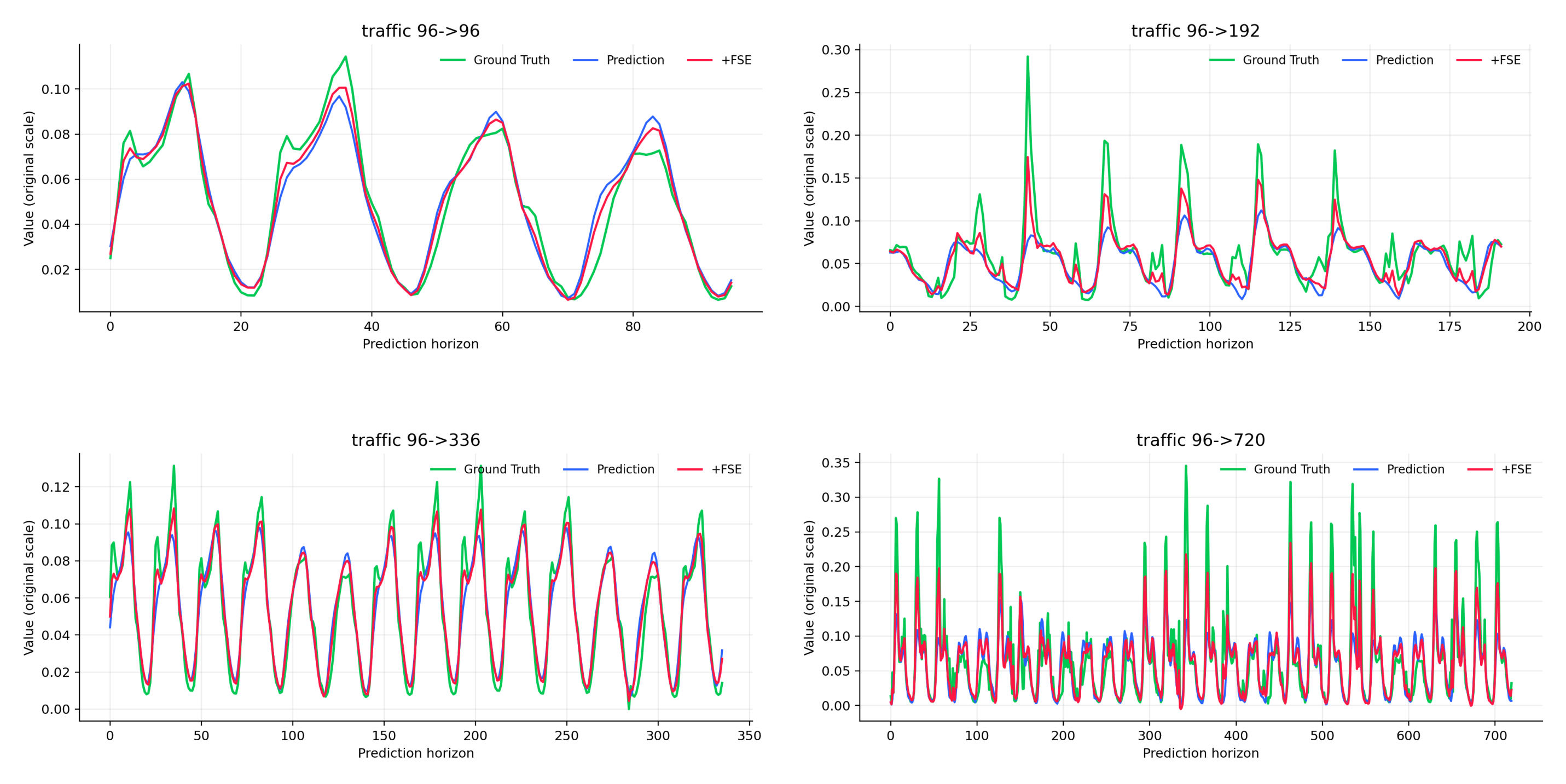}
    \caption{Time-domain forecasting visualization on the Traffic dataset across different prediction lengths.}
    \label{fig:tra_time}
\end{figure*}

Figures~\ref{fig:ecl_time} and~\ref{fig:tra_time} show the corresponding time-domain forecasts. Compared with the baseline, FSE better preserves local peaks, valleys, and rapid variations while maintaining the main temporal trend. The improvement is visible across both short and long prediction horizons, suggesting that future-distilled spectral guidance helps recover predictable fine-grained dynamics that are weakened in the original forecast.

\end{document}